\documentclass{article}
\usepackage{iclr2027_conference,times}
\usepackage{amsmath,amssymb,amsfonts}
\usepackage{graphicx}
\usepackage{booktabs}
\usepackage{caption}
\usepackage[section]{placeins}        %
\usepackage{xcolor}
\usepackage[hidelinks]{hyperref}
\usepackage{url}
\usepackage{tikz}
\usetikzlibrary{arrows.meta,positioning,shapes.geometric,fit}
\definecolor{figblue}{HTML}{DAE8FC}\definecolor{figblueln}{HTML}{6C8EBF}
\definecolor{figgray}{HTML}{F2F2F0}\definecolor{figgrayln}{HTML}{9A9992}
\definecolor{figorange}{HTML}{FFE6CC}\definecolor{figorangeln}{HTML}{D79B00}
\definecolor{figpurple}{HTML}{E1D5E7}\definecolor{figpurpleln}{HTML}{9673A6}
\definecolor{figgreen}{HTML}{D5E8D4}\definecolor{figgreenln}{HTML}{82B366}
\definecolor{figred}{HTML}{F8CECC}\definecolor{figredln}{HTML}{B85450}

\newcommand{\anli}{$\alpha$NLI}
\newcommand{\bz}{\ensuremath{\mathrm{B}_0}}
\newcommand{\dsuff}{\ensuremath{\mathbf{d}}}
\newcommand{\pp}{\,\textrm{pp}}
\newcommand{\rnorm}[1]{\lVert #1\rVert}

\title{You Only Pass Once:\\
Answering and Abstaining Together\\
in a Single Forward Pass\\
of a Frozen Language Model}

\newif\ifarxiv
\arxivtrue
\ifarxiv
  \iclrfinalcopy
  \newcommand{\yopoemail}[1]{\normalfont\texttt{#1}}
  \author{%
    \hspace*{-\tabcolsep}%
    \begin{tabular}[t]{@{}p{0.5\textwidth}@{}p{\dimexpr0.5\textwidth-2\tabcolsep\relax}@{}}
      Ziyang Luo\thanks{Equal contribution.} & Zhongyao Chu$^{*}$ \\
      \yopoemail{zluo352@gatech.edu} & \yopoemail{justedwardchu@gmail.com} \\[6pt]
      Xinjie He & Youting Wang \\
      \yopoemail{xh2442@columbia.edu} & \yopoemail{ginkoin613@gmail.com} \\[6pt]
      Xukui Qin & Runxiong Wu \\
      \yopoemail{kuschqin@gmail.com} & \yopoemail{calvin.wu@wisc.edu} \\[6pt]
      Yan-Syuan Chen & \\
      \yopoemail{yansyuanchen@utexas.edu} &
    \end{tabular}}
\else
\author{Anonymous authors\\
Paper under double-blind review}
\fi
\newcommand{\NLOSS}{19}\newcommand{\NBEAT}{31}\newcommand{\NCELLS}{36}\newcommand{\NPAR}{34}\newcommand{\NABOVE}{17}

\begin{document}
\maketitle
\ifarxiv\lhead{Preprint. Under review.}\fi %

\begin{abstract}
A frozen language model under-uses the evidence in its hidden states and answers confidently
when the context contains no answer. We fix both on the residual stream without changing any
weight. For accuracy, a small trained network (the probe) adds an input-dependent vector to
the residual at several middle layers; a removable hook, it raises Qwen2.5-1.5B on \anli{}
(abductive reasoning) from $51$\% to $73$\% accuracy. For abstention, a fixed vector (the
direction), the mean hidden state of contexts that contain the answer minus that of contexts
that do not, scores the hidden state; a low score abstains. On four QA datasets with human
answerability labels (Qwen2.5-7B), moving the direction to a dataset it was not fit on costs
under $3$ percentage points of decision accuracy (from $0.83$ in domain) on average and still beats
asking the model whether its context suffices; the same judgment trained as a classifier into
the probe's activation edit or into LoRA weights loses $16$ and $22$ points and ends below
asking. We set out to put both functions on a single forward pass, and found that the probe's
edit changes the very hidden state the direction scores, degrading it. A small map --- trained
without labels, only to reconstruct the unedited hidden state from the edited one --- removes
$63$\% of that interference on 1.5B. Refit on the hidden-state pairs of LoFiT or LoReFT, the
same recipe repairs those methods too: the steering module is replaceable. The resulting
system, YOPO, answers and abstains in one forward pass. On \anli{} it raises three-way
accuracy (correct answer when the context suffices, abstention when it does not) from $0.38$
to about $0.75$ on 1.5B, and stays within $0.02$ of a two-pass reference, which scores a
separate probe-off pass, on 1.5B, 3B and 7B, at half the cost. We also observe improvements on
the four QA datasets, where answerability is human-labeled and answers are generated as
free-form text. Near-parity with the two-pass reference holds across nine backbones spanning
six model families without backbone-specific hyperparameter tuning.
\end{abstract}

\section{Introduction}
\label{sec:intro}
Ask a frozen language model to reason over a context: choose the more plausible of two
explanations for a pair of observations \citep{bhagavatula2020}, or answer a question from a
document. Two failures dominate. The model under-uses what it has: its hidden states contain
evidence that never reaches the answer. And it over-trusts what it lacks: when the context has
no answer, it answers anyway. Both fixes live on the residual stream, so we build both there
and never change a backbone weight.

\textbf{Editing the residual stream raises accuracy.} We call any component that edits the
residual stream of a frozen model a \emph{writer}. Ours, the \emph{probe}, is a small MLP
that at three middle layers adds to the residual a vector whose direction and length depend
on the input. It raises \anli{} accuracy
from $50.8\%$ to $72.5\%$ on the 1.5B model (Qwen2.5-1.5B; \citealp{qwen2025qwen25}) and from $77.5\%$ to $82.4\%$ on 7B
(\S\ref{sec:writer}).
Trained on the same data, LoFiT \citep{yin2024lofit} (fixed per-head offset vectors) and
LoReFT \citep{wu2024reft} (a learned low-rank rewrite of a residual subspace) match or beat it
on its training task on 1.5B, and off that task it can lower accuracy; we use it because it is removable and because every later result holds for the other writers too.

\textbf{Scoring the residual stream enables abstention, and a fixed vector transfers best.} Prompting Qwen2.5-7B to say whether its context is sufficient reaches $0.713$ decision
accuracy (the fraction of contexts correctly judged sufficient or not) averaged over SQuAD~2.0,
RepLiQA, MuSiQue-Curated and MuSiQue-Full. A fixed vector $\dsuff$, the mean hidden state of sufficient examples minus that of
insufficient ones, reaches $0.825$ on the same forward pass. Following \citet{lavi2025} we use this \emph{direction} on reasoning tasks. Averaged over the twelve source-to-target pairs of the four
datasets, the direction loses $2.3$ decision-accuracy points from its in-domain $0.825$ and
still beats the target's own prompt by $8.9$; the same
judgment trained into the probe loses $16.4$ and trained into LoRA \citep{hu2022lora} weights $22.2$, ending $3.4$ and $2.3$ points below that prompt (\S\ref{sec:reader}).

\textbf{Running both on one forward pass creates a conflict on the smaller models.} Separate passes for the probe and the direction double inference cost. On one pass the
probe's edit changes the hidden state the direction scores: on 1.5B the direction's AUROC on
HellaSwag, a dataset it was not fit on, drops from $0.919$ to $0.836$. Training the judgment on edited
states would lose its transfer (\S\ref{sec:reader}), so we undo the edit before scoring: a
small \emph{map} is trained to reconstruct the unedited residual from the edited one, with a
squared-error loss and no label. On 1.5B it recovers $63\%$ of the lost AUROC ($0.836$ to $0.888$, probe off $0.919$); refit on LoFiT, LoReFT or LoRA pairs, it repairs those writers as well (\S\ref{sec:results}). The resulting system, \textbf{YOPO}
(You Only Pass Once), answers and abstains in one forward pass of a frozen backbone.

\textbf{Contributions.}
\begin{itemize}
\item[(i)] \emph{A writer abstraction and one instance.} Any edit to the residual stream is a
writer. Our probe raises \anli{} accuracy from $50.8\%$ to $72.5\%$ on 1.5B and from $77.5\%$ to
$82.4\%$ on 7B with no weight change; on the same data LoFiT, LoReFT and LoRA \citep{hu2022lora} match or exceed it on its training task on 1.5B
(\S\ref{sec:writer}, App.~\ref{app:steergrid}--\ref{app:lofit}).
\item[(ii)] \emph{The sufficiency judgment transfers across datasets only as a fixed vector.}
On four QA datasets with human labels (Qwen2.5-7B), the direction loses $2.3$ decision-accuracy
points cross-domain from its in-domain $0.825$; the same judgment trained into the probe loses $16.4$ and into LoRA weights $22.2$, and both end below the
target's own prompt (\S\ref{sec:reader}).
\item[(iii)] \emph{A label-free map that removes the probe's effect on the direction, and
works for other writers.} Trained only to reconstruct the unedited residual, it recovers $63\%$ / $54\%$ of the HellaSwag AUROC the probe costs the direction on 1.5B / 3B, returns the 7B AUROC to within $0.003$ of the unedited pass, and, refit on LoFiT, LoReFT or LoRA pairs, recovers $77\%$ to all of their in-domain loss
(\S\ref{sec:method}--\ref{sec:results}).
\item[(iv)] \emph{YOPO.} Probe, map and direction on one forward pass; nothing is trained on
sufficiency labels, and no label from the evaluated dataset is used. On \anli{} it doubles three-way accuracy (correct answer when the
context suffices, abstention when it does not) over the frozen 1.5B model ($0.375$ to $0.753$;
$0.754\pm0.016$ over four seeds), is within $0.02$ of a two-pass reference in domain on 1.5B, 3B
and 7B at half the cost (and on nine of ten backbones from six families, App.~\ref{app:family}),
and on four QA datasets with free-form answers, run end to end with three writers at 1.5B,
3B and 7B, it beats the frozen model in \NBEAT{} of \NCELLS{} (scale, writer, dataset) settings and is within $0.03$ of the two-pass reference in \NPAR{} (\S\ref{sec:benchmarks}, \S\ref{sec:nativelabel}).
\end{itemize}

\section{Setup}
\label{sec:setup}
\textbf{Names.} The \emph{residual stream} is the hidden-state vector each transformer layer
adds to and passes on. The \emph{probe} is our writer. The \emph{direction} $\dsuff$ is the
fixed scoring vector and the \emph{score} is $s=\dsuff^{\top}\mathbf{h}$ at the position before
the answer. The \emph{threshold} is the score below which the system abstains. The \emph{map} $M$ undoes the probe's edit
before scoring. \emph{YOPO} is probe, map, direction and threshold on one forward pass; the
\emph{two-pass reference} takes the answer from a pass with the probe on and the score from a
second pass with it off. \emph{In domain}: direction and threshold fit on the evaluated
dataset; \emph{cross-domain}: fit on a different one. ``1.5B'' means Qwen2.5-1.5B.

\textbf{The probe.} At three layers, about $0.4$, $0.55$ and $0.7$ of depth (the rule used on all but three
backbones; App.~\ref{app:data}), a small MLP reads the residual $\mathbf{r}$ at five prompt positions (the last token and
the tokens at $1/4$, $1/2$, $3/4$ and $7/8$ of the prompt) and replaces it with $\mathbf{r} + a\,\rnorm{\mathbf{r}}\,\mathbf{u}(\mathbf{r})$, with a unit
vector $\mathbf{u}(\mathbf{r})$ and a scalar $a$ both produced by the MLP. Only the MLP is
trained, with cross-entropy on the answer ($3.6$M parameters on 1.5B, $8.3$M on 7B); it is a
removable forward hook.

\textbf{The direction.} From labelled source examples, $\dsuff = \mathrm{mean}(\mathbf{h}\,|\,
\text{sufficient}) - \mathrm{mean}(\mathbf{h}\,|\,\text{insufficient})$ at one middle layer,
the \emph{read layer}, normalised to unit length. Nothing is optimised: the two means are
computed once. On the four QA datasets the threshold is
set on a held-out split used only for that purpose in domain and as the median score of unlabelled target inputs cross-domain; on the \anli{} benchmark of \S\ref{sec:yopo} (defined there) it is a percentile of the evaluation set's own scores. Cross-domain, no label of the evaluated dataset is used.

\textbf{Data and metrics.} The probe is measured by answer accuracy on \anli{}
\citep{bhagavatula2020}, HellaSwag \citep{zellers2019hellaswag} and PIQA \citep{bisk2020piqa}.
The direction is measured where human answerability labels exist: SQuAD~2.0
\citep{rajpurkar2018}, RepLiQA \citep{monteiro2024repliqa} and MuSiQue
\citep{trivedi2022musique} in a curated and a full version (App.~\ref{app:data}). AUROC measures how well the score ranks sufficient above insufficient contexts, independent of any threshold; decision accuracy is the fraction judged correctly once the threshold is applied.

\section{The Writer}
\label{sec:writer}
\textbf{Architecture.} At three middle layers of a frozen backbone, a small MLP
(the probe; Figure~\ref{fig:steerarch}) reads the residual at five prompt positions (\S\ref{sec:setup}) and
writes back $\mathbf{r}' = \mathbf{r} + a\,\rnorm{\mathbf{r}}\,\mathbf{u}(\mathbf{r})$, with
a direction and a length that both depend on the input. Only the probe is trained
($0.23\%$ of backbone parameters on 1.5B, answer cross-entropy); the edit is applied by a hook and removed by deleting it, and
no backbone weight changes. The probe's layers are $\{12,16,20\}$ on 1.5B/7B and $\{16,20,24\}$
on 3B.
Table~\ref{tab:writersets} (App.~\ref{app:data}) lists, for every comparison in this section,
the backbone, the training set and the evaluation set; numbers from different rows of that table use different data and are not comparable.
\suppressfloats[t]
\begin{figure}[h]\centering
\resizebox{0.72\textwidth}{!}{\begin{minipage}{0.72\textwidth}
\resizebox{\linewidth}{!}{\begin{tikzpicture}[font=\scriptsize, >={Stealth[length=1.6mm]}, node distance=2.2mm,
  blk/.style={draw, rounded corners=2.5pt, minimum width=31mm, minimum height=5.2mm, align=center, line width=0.7pt},
  frz/.style={blk, fill=figgray, draw=figgrayln},
  inj/.style={blk, fill=figblue, draw=figblueln, font=\scriptsize\bfseries},
  io/.style={blk, fill=figgreen, draw=figgreenln},
  ours/.style={blk, fill=figorange, draw=figorangeln}]
  \node[io] (emb) {Embedding};
  \node[frz, above=of emb]   (la) {Layers 0--11 (frozen)};
  \node[inj, above=of la]    (l12) {Layer 12};
  \node[frz, above=of l12]   (lb) {Layers 13--15 (frozen)};
  \node[inj, above=of lb]    (l16) {Layer 16};
  \node[frz, above=of l16]   (lc) {Layers 17--19 (frozen)};
  \node[inj, above=of lc]    (l20) {Layer 20};
  \node[frz, above=of l20]   (ld) {Layers 21--27 (frozen)};
  \node[io, above=of ld]     (head) {RMSNorm + LM head};
  \foreach \a/\b in {emb/la, la/l12, l12/lb, lb/l16, l16/lc, lc/l20, l20/ld, ld/head} \draw[->] (\a) -- (\b);
  \node[below=2mm of emb, font=\scriptsize\itshape] (tok) {input tokens};   \draw[->] (tok) -- (emb);
  \node[above=2mm of head, font=\scriptsize\itshape] (out) {answer logits}; \draw[->] (head) -- (out);
  \node[blk, fill=figblue, draw=figblueln, minimum width=46mm, right=26mm of l20, anchor=north west, xshift=-2mm] (read)
    {read $\mathbf r$ at spread-5 positions ($B\times5\times1536$)};
  \node[ours, minimum width=22mm, below=2.6mm of read.south west, anchor=north west] (dir)
    {direction MLP\\$1536{\to}384{\to}1536$\\SiLU $\to\ \mathbf u(\mathbf r)$};
  \node[ours, minimum width=22mm, below=2.6mm of read.south east, anchor=north east] (gat)
    {learned magnitude\\$1536{\to}5$,\ $\sigma\times\alpha_{\max}$\\$\to\ a$};
  \node[ours, minimum width=46mm, below=2.6mm of dir.south west, anchor=north west, font=\scriptsize\bfseries] (wr)
    {write\ \ $\mathbf r \leftarrow \mathbf r + a\,\rnorm{\mathbf r}\,\mathbf u(\mathbf r)$};
  \draw[->] (read.south -| dir) -- (dir.north -| dir);
  \draw[->] (read.south -| gat) -- (gat.north -| gat);
  \draw[->] (dir.south) -- (wr.north -| dir.south);
  \draw[->] (gat.south) -- (wr.north -| gat.south);
  \node[draw=figorangeln, dashed, rounded corners=4pt, inner sep=2.6mm, line width=0.7pt, fit=(read)(dir)(gat)(wr),
        label={[figorangeln, font=\scriptsize\bfseries]above:steering sub-probe ($\times3$, identical design)}] (probe) {};
  \foreach \l/\dx in {l20/3mm, l16/5.5mm, l12/8mm}
    \draw[figblueln, dashed, ->] ([yshift=1.4mm]\l.east) -- ++(\dx,0) |- ([yshift=1mm]read.west);
  \foreach \l/\dx in {l20/-9mm, l16/-11mm, l12/-13mm}
    \draw[figorangeln, line width=0.9pt, ->] (wr.west) -- ++(\dx,0) |- ([yshift=-1.4mm]\l.east);
  \node[below=5mm of wr, font=\tiny\itshape, align=center, text=black!55]
    {$\approx$3.58M params ($0.23\%$ of backbone) $\cdot$ probe-only training (answer cross-entropy)\\[1pt]
     dashed = read (hook)\quad solid = write-back\quad blue = injection layers\quad gray = frozen};
\end{tikzpicture}}
\end{minipage}}
\caption{The probe. At each injection layer a small MLP reads the residual at five prompt
positions and writes back an input-dependent vector through a forward hook; only the probe is
trained, the backbone is frozen. Shown: the 1.5B instantiation (injection layers 12/16/20).}
\label{fig:steerarch}
\end{figure}

\textbf{Accuracy gain.} On \anli{} the probe lifts a frozen Qwen2.5-1.5B from $50.78$
to $72.52$ ($+21.74$, exact McNemar $p{=}2.7{\times}10^{-36}$) and the 7B model from $77.48$ to $82.44$
($+4.96$, $p{=}1.7{\times}10^{-7}$), both with 15k training items and 3 epochs (the grid of
App.~\ref{app:steergrid} uses 5k items and 1 epoch and shows smaller gains; frozen baselines
in App.~\ref{app:steergrid}--\ref{app:lofit} come from separate runs and differ from the
values here by up to $0.5$ points). The gain is
largest where the frozen model is weakest: on \anli{}, $49.9\to61.0$ on Qwen2.5-0.5B ($+11.2$),
$67.0\to77.7$ on 3B ($+10.8$), $74.6\to77.6$ on Phi-3-mini (\citealp{abdin2024phi3}; $+2.9$) and $73.7\to73.8$ on
Mistral-7B (\citealp{jiang2023mistral}; $+0.07$; 5k items, 1 epoch). Swapping the two answer options changes the frozen
model's prediction on $5\%$ of items and the probe-on model's on $78\%$: the probe reads the
options, not their position.

\textbf{Comparisons on the same training data.} With the same training data, the probe is not the most accurate writer
on its training task (Table~\ref{tab:matched}, App.~\ref{app:lofit}); what distinguishes it is that the edit is an
input-conditioned function of the residual it reads, at no change to any weight. LoFiT
\citep{yin2024lofit}, the nearest neighbour, static per-head offset vectors with $4.4$k
trainable parameters at 1.5B, matches the probe on its training task on 1.5B and exceeds it on 3B/7B with
$10^2$--$10^3\times$ fewer parameters; the full three-scale comparison with pre-registered
equal-budget and matched-cost controls is in App.~\ref{app:lofit}. The consequences that
matter are system-level (\S\ref{sec:results}): on 1.5B every writer we tested lowers the
direction's AUROC, and the same label-free map recipe repairs each of them.

\textbf{On a dataset it was not trained on, the probe can lower accuracy.} Trained on \anli{} and evaluated
zero-shot, the probe helps on PIQA ($68.7\to72.9$, $p{=}0.002$) and lowers HellaSwag ($73.6\to69.9$;
zero-shot, 1,500 items; the App.~\ref{app:steergrid} grid evaluates a different 1,532-item
HellaSwag split, frozen $74.1$); our reimplementation of a LoReFT-style edit (App.~\ref{app:lofit}) also lowers HellaSwag
($-1.87$, $p{=}0.079$, not significant), while LoRA \citep{hu2022lora} raises it ($+5.93$). Against our reimplementation of LoFiT and the official LoReFT code, transfer results are mixed: the probe transfers better on 7B and on BoolQ at 1.5B, worse on HellaSwag at 1.5B (App.~\ref{app:lofit}).

\section{The Reader}
\label{sec:reader}
When the context itself lacks the information needed to answer, no amount of reasoning can
recover the answer. The first question is therefore not whether the model answers correctly,
but whether it should answer at all. The methods \S\ref{sec:writer} uses to improve reasoning
apply to this question in principle, so we ask: are they still of any use here? To find out, we
retrain the probe and LoRA of \S\ref{sec:writer} directly on this judgment (same architecture, only
the training target changes) and set them against two systems that are not trained: a
read-only difference-of-means direction \citep{lavi2025}, and a zero-shot baseline that prompts
the model to answer directly. The comparison runs on four QA domains, and every answerability
label comes with its dataset.

All four systems share one prompt, unchanged across domains: after the passage and the question
it asks \emph{Does the passage contain enough information to answer the question? Answer 1
(enough) or 2 (not enough)}. The zero-shot baseline and both writers decide by the argmax of the
two option logits; the reader reads the residual stream at the last token of the same prompt and
decides with a threshold. All four are scored by decision accuracy on the same test items.

\textbf{Data construction.} Each domain has a development pool of 2,000 items, exactly 1:1
answerable to unanswerable, split with stratification into \emph{fit} (1,200) and \emph{sel}
(800), and a disjoint \emph{test} set of 2,000 items (1,000/1,000). Fit is used only to compute
the direction, sel only to select the layer and set the threshold, and test only to report.

\textbf{Constructing the reader.} The reader (``read direction'' in the tables) is the direction of \S\ref{sec:setup}, refit here: the difference between the mean residual stream of unanswerable
and of answerable examples at one layer. It has no free parameters; the direction is a
closed-form statistic, not a fitted model. Concretely, at every layer $\ell$ we take the residual
stream $h$ at the last token and compute, on fit, $d_\ell = \mathrm{mean}(h\,|\,\text{unanswerable})
- \mathrm{mean}(h\,|\,\text{answerable})$ (the opposite orientation to the $\dsuff$ of
\S\ref{sec:setup}; the two differ only by a sign, and the decisions are identical).

The reader selects its layer without any target-domain label. On the source domain's sel, we
inject $d_\ell$ at the last-token position, layer by layer, record each item's steering response, the log-odds of answer ``2'' against all other tokens, $\mathrm{logit}(\text{``2''}) - \mathrm{logsumexp}(\text{all other logits})$, and average over all items to obtain $\psi(\ell)$; this step uses no label at all. The read layer is where $\psi$ rises fastest from one layer to the next: $L_0 = \arg\max_\ell\,[\psi(\ell) - \psi(\ell-1)]$.

The decision is a scalar threshold. Let $\hat d$ be the unit vector of the direction at layer
$L_0$ and $h$ the residual stream at the last token of that layer. The score is $s = \hat d \cdot h$,
and $s > \theta$ is judged unanswerable. In domain, $\theta$ is the median of $s$ over the
domain's own sel; cross-domain, only the threshold changes: the direction still comes from the
source domain, and $\theta$ becomes the median over the target domain's own sel. The entire cost
of cross-domain transfer is this one scalar, and it needs no label. The median can serve directly
as the threshold only because the classes are exactly 1:1, which the data construction
guarantees.

\textbf{Constructing the two writers.} The probe is the activation writer
of \S\ref{sec:writer}, and LoRA is the $r{=}3$ low-rank adapter of \S\ref{sec:writer}. Neither
architecture is changed in any way; only the training target switches from the original task to
this judgment. The training set is fit + sel: each domain's 2,000 items, exactly the union of the
fit 1,200 and the sel 800 that the reader uses separately, are used for training as a whole.

\textbf{Comparison setup.} Four backbones: Qwen2.5-1.5B / 3B / 7B-Instruct and
Mistral-7B-Instruct-v0.3. For a writer, cross-domain means the checkpoint trained on the source
domain is evaluated on the target's test set without any adjustment; for the reader, the source domain's direction with a threshold at the median score on the target domain (no label needed). Table~\ref{tab:reader7b}
reports Qwen2.5-7B; App.~\ref{app:readerscales} the other three backbones.

\ifarxiv\suppressfloats[t]\begin{table}[h]\else\begin{table}[t]\fi\centering\scriptsize
\setlength{\tabcolsep}{2pt}
\caption{In-domain and cross-domain decision accuracy on Qwen2.5-7B-Instruct (2,000 test
items). Rows are source domains, columns target domains; in-domain cells in bold. Writer cells
average three training seeds. Zero-shot baseline (no source domain; shared by all three groups):
SQuAD2 .856, RepLiQA .870, MQ-C .589, MQ-F .535.}
\label{tab:reader7b}
\begin{tabular}{l cccc cccc cccc}
\toprule
& \multicolumn{4}{c}{read direction} & \multicolumn{4}{c}{probe} & \multicolumn{4}{c}{LoRA $r{=}3$}\\
\cmidrule(lr){2-5}\cmidrule(lr){6-9}\cmidrule(lr){10-13}
source $\backslash$ target & SQuAD2 & RepLiQA & MQ-C & MQ-F & SQuAD2 & RepLiQA & MQ-C & MQ-F & SQuAD2 & RepLiQA & MQ-C & MQ-F\\
\midrule
SQuAD2  & \textbf{.856} & .906 & .764 & .687 & \textbf{.854} & .833 & .623 & .555 & \textbf{.892} & .756 & .694 & .591\\
RepLiQA & .837 & \textbf{.952} & .744 & .689 & .819 & \textbf{.975} & .588 & .545 & .694 & \textbf{.997} & .622 & .565\\
MQ-C    & .850 & .894 & \textbf{.793} & .704 & .611 & .722 & \textbf{.847} & .618 & .677 & .674 & \textbf{.942} & .569\\
MQ-F    & .852 & .913 & .779 & \textbf{.700} & .750 & .722 & .752 & \textbf{.694} & .829 & .786 & .813 & \textbf{.816}\\
\midrule
in-domain mean & \multicolumn{4}{c}{.825} & \multicolumn{4}{c}{.842} & \multicolumn{4}{c}{.912}\\
cross-domain mean & \multicolumn{4}{c}{.802} & \multicolumn{4}{c}{.678} & \multicolumn{4}{c}{.689}\\
\begin{tabular}[t]{@{}l@{}}cross-domain $-$\\in-domain (points)\end{tabular} & \multicolumn{4}{c}{$-2.3$} & \multicolumn{4}{c}{$-16.4$} & \multicolumn{4}{c}{$-22.2$}\\
\begin{tabular}[t]{@{}l@{}}cross-domain $-$ prompting\\baseline (points)\end{tabular} & \multicolumn{4}{c}{$+8.9$} & \multicolumn{4}{c}{$-3.4$} & \multicolumn{4}{c}{$-2.3$}\\
\bottomrule
\end{tabular}
\end{table}

\textbf{The ranking reverses between in domain and cross-domain.} On Qwen2.5-7B-Instruct
(Table~\ref{tab:reader7b}), both writers do well in domain: the probe is $13.0$ points
above the zero-shot baseline on average and LoRA $19.9$, the latter ranking first on all four
domains. Cross-domain, both collapse: they drop $16.4$ and $22.2$ points, to $.678$ and $.689$,
which is $3.4$ and $2.3$ points below the target domain's own zero-shot; on this backbone,
carrying a trained judgment to a new domain is on average worse than carrying nothing. The
read-only difference-of-means direction, by contrast, is $11.3$ points above the baseline in
domain and still $8.9$ above it cross-domain, falling only from $.825$ to $.802$, a drop of $2.3$
points. The same reversal replicates in another model family (Mistral-7B-Instruct-v0.3) and at
1.5B and 3B (App.~\ref{app:readerscales}).

\textbf{Summary.} On all four backbones the two writers are $11.3$--$24.6$ points above the
baseline in domain and fall to about the baseline or below cross-domain (the probe $1.6$--$3.4$
below it, LoRA between $-2.3$ and $+2.7$), whereas the read direction beats the baseline
cross-domain by $8.8$--$12.5$ points and drops only $2.1$--$3.4$ against $13.5$--$24.6$ for the
writers. The judgment of whether to answer should therefore stay on the read side.

\section{YOPO: One Pass}
\label{sec:yopo}
Run separately, the probe and the direction cost two forward passes per query, one with the
probe on to answer and one with it off to score. This section puts them on one pass.

\textbf{The four-condition benchmark.} Each \anli{} item is presented in four versions: full
(both observations; sufficient), rm-obs1 and rm-obs2 (one observation removed; still answerable), replace-both
(both replaced by unrelated text; insufficient); HellaSwag likewise. AUROC is computed on full vs.\ replace-both; three-way accuracy (\S\ref{sec:benchmarks}) on all four. Nothing is fit on HellaSwag; every fixed setting is in App.~\ref{app:data}.

\subsection{The Problem: on the Smaller Models the Probe's Edit Lowers the Direction's AUROC}
\label{sec:problem}
When the direction scores a pass the probe has edited, it separates sufficient from insufficient contexts less well. On 1.5B the in-domain AUROC falls from $0.962$ on the pass with the probe off
to $0.913$ on the pass with it on, and the HellaSwag AUROC from $0.919$ to $0.836$. The loss
shrinks with model size: on HellaSwag it is $0.083$, $0.031$ and $0.016$ on 1.5B, 3B and 7B,
and on 7B the edited pass already scores better in domain ($0.995$ vs.\ $0.985$). The loss is not specific to our construction or our probe: on SQuAD~2.0 with human labels
the same edit lowers the AUROC from $0.836$ to $0.782$ on 1.5B (\S\ref{sec:nativelabel}), and
LoFiT and LoReFT cost more on 1.5B (\S\ref{sec:results}). Training a sufficiency classifier on the edited states would fix this in domain but not transfer (\S\ref{sec:reader}); we undo the edit instead.

\subsection{Method: Undo the Edit Before Scoring}
\label{sec:method}
The probe's edit depends on the input but not on whether the context is sufficient. A network
trained to reconstruct the unedited residual from the edited one therefore learns to undo the
edit, not to judge sufficiency:
\begin{equation}
\min_M \;\mathbb{E}\,\big\lVert M\!\left(h^{\mathrm{on}}\right) - h^{\mathrm{off}}\big\rVert_2^2,
\qquad s = \dsuff^{\top} M\!\left(h^{\mathrm{on}}\right),
\end{equation}
where $h^{\mathrm{off}}$ and $h^{\mathrm{on}}$ are the residuals at the read layer (layer 18
on 1.5B and 7B, 24 on 3B) for the same input with the probe off and on. The pairs cost one
extra forward pass per training item and no annotation; $\dsuff$ is not changed. The map is
$M(h)=h+U\,\sigma(Vh)$ with rank 64 ($V$: $d\to64$, SiLU, $U$: $64\to d$), $U$ initialised at zero and $V$ at PyTorch's
default, so $M$ starts exactly at the identity and learns only the correction (the pre-registered recipe started $U$ at random; its results are in App.~\ref{app:failures}, item (x)). Rank, optimiser and steps were fixed before any run and are identical at every scale and backbone (App.~\ref{app:data}).

At inference YOPO runs one pass with the probe on, computes $s=\dsuff^{\top}M(h^{\mathrm{on}})$
at the read layer, abstains if $s$ is below the threshold, and otherwise returns the answer of
that same pass (Figure~\ref{fig:arch}). Nothing is trained on sufficiency labels: the probe is trained on answers, $\dsuff$ is the
two class means of labelled source examples, $M$ is trained on residual pairs; no target label
is used.
\suppressfloats[t]
\begin{figure}[h]\centering
\resizebox{\textwidth}{!}{%
\begin{tikzpicture}[font=\small, >={Stealth[length=2mm]},
  box/.style={draw, rounded corners=2pt, minimum height=7mm, inner sep=4pt, align=center},
  lay/.style={draw=gray!70, fill=gray!12, minimum height=8mm, inner sep=0pt}]
  \node[box, fill=figgreen, draw=figgreenln] (inp) at (0,0) {Context\\+ Question};
  \node[lay, draw=figgrayln, fill=figgray, minimum width=26mm, right=5mm of inp,
        label={[gray,font=\scriptsize]below:{$L_0\ldots L_{11}$ (probe off)}}] (early) {};
  \node[lay, fill=figblue, draw=figblueln, minimum width=44mm, right=0mm of early,
        label={[figblueln,font=\scriptsize,xshift=-10mm]below:{$L_{12}\ldots L_{27}$ (probe on)}}] (late) {};
  \node[box, fill=figgreen, draw=figgreenln, right=5mm of late] (logits) {answer\\logits};
  \draw[->] (inp) -- (early);
  \draw[->] (late) -- (logits);
  \node[gray, font=\itshape\scriptsize, below=4mm of early.south] {frozen LLM: residual stream};
  \node[box, fill=figblue, draw=figblueln, below=13mm of late.center, xshift=6mm] (steer) {the probe\\\scriptsize adds $a\rnorm{\mathbf{r}}\mathbf{u}(\mathbf{r})$ at layers $\{12,16,20\}$};
  \draw[->, figblueln, thick] (steer) -- node[right, font=\scriptsize, pos=0.85]{$\oplus$} (late.south -| steer);
  \node[box, fill=figorange, draw=figorangeln, dashed, above=12mm of early.east]
    (dec) {map $M$\\\scriptsize label-free: trained probe-on$\to$probe-off, $\dsuff$ fixed};
  \coordinate (readpt) at ([xshift=5mm]late.north west);
  \draw[->, figpurpleln, dashed, thick] (readpt) -- node[right, font=\scriptsize, figpurpleln]{$h^{\mathrm{on}}$ at layer 18} (dec.south -| readpt);
  \node[box, fill=figpurple, draw=figpurpleln, right=7mm of dec] (gate) {score $s=\dsuff^{\top} M(h^{\mathrm{on}})$};
  \draw[->] (dec) -- (gate);
  \node[diamond, draw, aspect=2.2, inner sep=1.5pt, right=6mm of gate, font=\scriptsize] (dec2) {$s<\tau$?};
  \draw[->] (gate) -- (dec2);
  \node[box, fill=figred, draw=figredln, right=6mm of dec2] (abst) {abstain (0)};
  \draw[->, figredln] (dec2) -- node[above, font=\scriptsize, figredln]{yes} (abst);
  \draw[->, figgreenln] (dec2.south) -- node[pos=0.35, right, font=\scriptsize, align=left]{else: that pass's\\answer (1/2)} (logits.north -| dec2.south) -- (logits.north);
\end{tikzpicture}}
\caption{YOPO on one forward pass. The probe (blue) edits the residual stream; the
same pass is read at the read layer, passed through the map $M$ (orange, trained only to
reconstruct the probe-off residual), and scored with the direction $\dsuff$ (purple).
Below the threshold $\tau$ the system abstains; otherwise it returns that pass's answer. Layer indices are for
Qwen2.5-1.5B/7B; 3B uses $\{16,20,24\}$ with read layer 24.}
\label{fig:arch}
\end{figure}

\subsection{Results: the Map Works at Three Scales and for Four Writers}
\label{sec:results}
\textbf{The map recovers over half of the loss without labels.} Probe rows of Table~\ref{tab:writers}: on 1.5B the map raises the direction's HellaSwag AUROC from $0.836$ to $0.888$ ($63\%$ of the loss) and its in-domain AUROC from $0.913$ to $0.944$; on 3B it recovers $54\%$ ($0.891\to0.908$, probe off $0.922$). On 7B
the probe raises the in-domain AUROC ($0.985$ to $0.995$) and lowers the HellaSwag AUROC by
$0.016$; the map returns both to the probe-off level ($0.986$ and $0.965$).

\textbf{The same map recipe, refit per writer, reduces each writer's loss.} The recipe uses nothing specific to our probe. Refit on the (on, off) pairs of another writer, it repairs that writer too
(Table~\ref{tab:writers}). LoFiT's fixed per-head offsets \citep{yin2024lofit} lower the
direction's in-domain AUROC by $0.178$ on 1.5B and $0.109$ on 3B, $3.6\times$ and $4.4\times$
the probe's loss; the refit map recovers $92\%$ and $77\%$ of it (Table~\ref{tab:writers}). LoReFT's subspace edit \citep{wu2024reft} lowers it by $0.232$ on 1.5B and
the map recovers all of
it ($0.730\to0.971$); on 3B the loss is $0.028$ and the map more than recovers it ($0.939\to0.975$). LoRA \citep{hu2022lora} at matched budget, a weight edit, lowers it by $0.151$ on 1.5B and the map recovers all of it ($0.811\to0.962$); on 3B LoRA does not lower the AUROC ($0.967$ off, $0.972$ on) and the map
leaves it there ($0.974$). On HellaSwag the refit map recovers $72$--$85\%$ of each writer's loss on 1.5B and $51\%$
(LoFiT) and $33\%$ (LoReFT) on 3B, where the losses are only $0.115$ and $0.063$; the one cell where the map costs
something is LoRA on 3B on HellaSwag ($0.921\to0.893$; Table~\ref{tab:writers}). End to end (three-way accuracy, defined in \S\ref{sec:benchmarks}), on 1.5B LoFiT, LoReFT
and LoRA with their refit maps reach $0.843$, $0.873$ and $0.888$ against $0.753$ for YOPO with the probe: a
stronger writer, once its effect on the direction is removed, gives a stronger system.

\ifarxiv\suppressfloats[t]\begin{table}[h]\else\begin{table}[t]\fi\centering\footnotesize
\setlength{\tabcolsep}{2.6pt}
\caption{The map at three scales and for four writers. AUROC of the same direction $\dsuff$
(full vs.\ replace-both) at the read layer on the pass with the writer off, the pass with it
on, and that pass after the map; in domain (\anli{}) and on HellaSwag (not fit on);
Qwen2.5-Instruct. The map is refit on each writer's own (on, off) \anli{}-train pairs and sees
no label. The writer-off pass is what the two-pass reference scores, at twice the cost. LoFiT
and LoReFT are the official-configuration writers of App.~\ref{app:lofit}; LoRA is the $r{=}3$
adapter of the matched-budget table in App.~\ref{app:lofit} (its 3B counterpart trained
identically).}
\label{tab:writers}
\begin{tabular}{ll ccc ccc}
\toprule
& & \multicolumn{3}{c}{\anli{} AUROC (in-domain)} & \multicolumn{3}{c}{HellaSwag AUROC (not fit on)}\\
\cmidrule(lr){3-5}\cmidrule(lr){6-8}
writer & scale & writer off & writer on & after map & writer off & writer on & after map\\
\midrule
probe (ours) & 1.5B & $0.962$ & $0.913$ & $\textbf{0.944}$ & $0.919$ & $0.836$ & $\textbf{0.888}$\\
probe (ours) & 3B & $0.967$ & $0.942$ & $\textbf{0.969}$ & $0.922$ & $0.891$ & $\textbf{0.908}$\\
probe (ours) & 7B & $0.985$ & $0.995$ & $\textbf{0.986}$ & $0.968$ & $0.952$ & $\textbf{0.965}$\\
\midrule
LoFiT & 1.5B & $0.962$ & $0.784$ & $0.947$ & $0.919$ & $0.751$ & $0.872$\\
LoFiT & 3B & $0.967$ & $0.858$ & $0.942$ & $0.922$ & $0.807$ & $0.866$\\
LoReFT & 1.5B & $0.962$ & $0.730$ & $0.971$ & $0.919$ & $0.738$ & $0.892$\\
LoReFT & 3B & $0.967$ & $0.939$ & $0.975$ & $0.922$ & $0.859$ & $0.880$\\
LoRA & 1.5B & $0.962$ & $0.811$ & $0.962$ & $0.919$ & $0.784$ & $0.888$\\
LoRA & 3B & $0.967$ & $0.972$ & $0.974$ & $0.922$ & $0.921$ & $0.893$\\
\bottomrule
\end{tabular}
\end{table}

\subsection{End to End: In Domain the Pair Beats Either Component}
\label{sec:benchmarks}
\textbf{Measurement.} \anli{} four-condition validation (400 items, 100 per version) and HellaSwag (500,
nothing fit on it); three-way accuracy (three outputs: option 1, option 2, abstain); threshold at the 25th percentile of the evaluation set's own scores ($p{=}25$), which matches the one-in-four share of insufficient versions; no label is used.
Table~\ref{tab:combined} reports seed 0; the seed spread is in Table~\ref{tab:combined_seeds} (App.~\ref{app:data}).

\begin{table}[t]\centering\footnotesize
\setlength{\tabcolsep}{4pt}
\caption{End-to-end three-way accuracy (correct answer on the three answerable versions,
abstention on replace-both), four-condition validation sets, seed 0; abstention threshold at the 25th percentile.}
\label{tab:combined}
\begin{tabular}{lcccccc}
\toprule
policy & \multicolumn{2}{c}{1.5B} & \multicolumn{2}{c}{3B} & \multicolumn{2}{c}{7B}\\
 & \anli{} & HellaSwag & \anli{} & HellaSwag & \anli{} & HellaSwag\\
\midrule
frozen model (never abstains) & $0.375$ & $0.500$ & $0.563$ & $0.400$ & $0.628$ & $0.604$\\
probe alone (never abstains) & $0.590$ & $0.502$ & $0.625$ & $0.502$ & $0.668$ & $0.592$\\
direction alone (probe off) & $0.560$ & $0.602$ & $0.743$ & $0.560$ & $0.828$ & $0.780$\\
two-pass reference & $0.753$ & $0.598$ & $0.790$ & $0.642$ & $0.863$ & $0.768$\\
one pass, no map & $0.698$ & $0.580$ & $0.760$ & $0.616$ & $0.880$ & $0.754$\\
\textbf{YOPO} (one pass) & $0.753$ & $0.606$ & $0.800$ & $0.626$ & $0.858$ & $0.768$\\
\bottomrule
\end{tabular}
\end{table}

\textbf{In domain, the pair beats each component and is within $0.02$ of two passes.} On 1.5B the frozen model scores $0.375$ (it always picks one option), the probe
alone $0.590$, the direction alone $0.560$ and YOPO $0.753$; 3B and 7B follow the same order
(Table~\ref{tab:combined}). YOPO matches the two-pass reference on 1.5B ($0.753$ vs.\ $0.753$), is $0.01$ above it on 3B ($0.800$ vs.\ $0.790$) and
$0.005$ below it on 7B ($0.858$ vs.\ $0.863$), at half the cost. Over seeds, YOPO vs.\ the two-pass reference: $0.754\pm0.016$
vs.\ $0.754\pm0.010$ on 1.5B, $0.805\pm0.004$ vs.\ $0.802\pm0.008$ on 3B, $0.868\pm0.007$ vs.\
$0.863\pm0.001$ on 7B; YOPO beats both components on every seed at every scale.

\textbf{On HellaSwag the probe does not help, and YOPO gains nothing over the direction
alone.} There the direction with the probe off scores $0.602$ and $0.780$ on 1.5B and 7B, against
$0.606$ and $0.768$ for YOPO and $0.598$ and $0.768$ for the two-pass reference; on 3B YOPO trails
the two-pass reference ($0.626$ vs.\ $0.642$). The probe, trained on \anli{}, gains nothing on HellaSwag at 1.5B and 7B and gains only at 3B ($0.502$ vs.\ $0.400$). Where the probe helps, one pass is as good as two; where it does not, the
right deployment is to turn the probe off.

\subsection{Human Answerability Labels}
\label{sec:nativelabel}
The four-condition sets create insufficiency by deleting text, and a difference-of-means
vector separates full from replace-both at layer 11 already (AUROC $0.905$), earlier than a
sufficiency judgment could form: the numbers above are results on a controlled construction.
This section
repeats two claims on human labels: the probe's edit lowers the direction's AUROC and the map
repairs it; and scoring the generation pass itself, with no second pass, matches a two-pass
reference.

\textbf{The edit lowers the AUROC and the map repairs it (1.5B, 3B).} We keep the \anli{}
probe unchanged, use each dataset's sufficiency question (the prompt of \S\ref{sec:reader}, the \emph{judgment prompt}) as input, and its human
answerable/unanswerable test labels as target. On SQuAD~2.0 the direction scores $0.836$ AUROC
with the probe off and $0.782$ with it on; the map, fit on 1,200 unlabelled items, restores
$0.811$, $54\%$ of the loss (both pre-registered hypotheses supported); on 3B
$0.872\to0.835$, and $0.856$ after the map ($56\%$). On RepLiQA the direction scores $0.99$ throughout: nothing to repair.

\textbf{YOPO end to end on four QA datasets (1.5B, 3B, 7B; three writers).} Here the writer
answers in free-form text. Each writer (the probe; LoFiT; LoReFT) is trained on the answerable
rows of the dataset's fit split with the deployed prompt and cross-entropy on the gold answer;
unanswerable rows are not used, so nothing about abstention is trained in. The map is refit on (writer-on, writer-off) hidden states of the fit
split at the read layer, the direction and the read layer are chosen on the generation prompt with the
writer off (two-means direction; the layer with the best AUROC on the labelled selection split, allowed
here because deployment is in domain), and
the threshold is set on the selection split with the writer on. At test time the writer is on
during generation; the direction scores that pass's hidden state at the last prompt token, after the map, and the
system abstains when the score is low or the generation itself says ``unanswerable''. The
two-pass reference keeps the writer-on generation and scores a separate judgment-prompt pass.
Table~\ref{tab:m51main} reports answer-or-abstain accuracy; per-cell answerable accuracy and
AUROC are in App.~\ref{app:suite}, Table~\ref{tab:m51interf}.

\begin{table}[t]\centering\scriptsize
\setlength{\tabcolsep}{2.4pt}
\caption{YOPO end to end on four QA datasets (Qwen2.5-Instruct, test split, 2,000 items per
dataset, answer-or-abstain accuracy). Each writer cell gives YOPO (writer on during generation;
map and direction on the same pass) and the two-pass reference (the same writer-on generation,
scored on a separate judgment-prompt pass). Bold: YOPO within $0.02$ of, or above, the two-pass
reference. The read layer is chosen on the deployed prompt (App.~\ref{app:suite}). Results with no writer (frozen model, direction only) are in App.~\ref{app:suite}, Table~\ref{tab:taskc_scales}.}
\label{tab:m51main}
\resizebox{\textwidth}{!}{%
\begin{tabular}{l l cccc}
\toprule
scale & system (answer-or-abstain accuracy: YOPO / two-pass reference) & SQuAD2 & RepLiQA & MuSiQue-C & MuSiQue-F\\
\midrule
1.5B & frozen model, generation alone & $0.591$ & $0.439$ & $0.560$ & $0.410$\\
1.5B & probe on: YOPO / two-pass & $\textbf{0.700}$ / $0.711$ & $\textbf{0.788}$ / $0.792$ & $\textbf{0.654}$ / $0.638$ & $\textbf{0.423}$ / $0.404$\\
1.5B & LoFiT on: YOPO / two-pass & $0.696$ / $0.720$ & $\textbf{0.804}$ / $0.805$ & $\textbf{0.637}$ / $0.635$ & $\textbf{0.424}$ / $0.408$\\
1.5B & LoReFT on: YOPO / two-pass & $0.693$ / $0.714$ & $\textbf{0.797}$ / $0.802$ & $\textbf{0.661}$ / $0.644$ & $\textbf{0.431}$ / $0.410$\\
\midrule
3B & frozen model, generation alone & $0.672$ & $0.602$ & $0.585$ & $0.483$\\
3B & probe on: YOPO / two-pass & $\textbf{0.759}$ / $0.775$ & $\textbf{0.842}$ / $0.853$ & $\textbf{0.682}$ / $0.696$ & $\textbf{0.467}$ / $0.452$\\
3B & LoFiT on: YOPO / two-pass & $\textbf{0.788}$ / $0.777$ & $\textbf{0.848}$ / $0.863$ & $0.667$ / $0.690$ & $0.419$ / $0.450$\\
3B & LoReFT on: YOPO / two-pass & $\textbf{0.777}$ / $0.777$ & $0.827$ / $0.849$ & $0.645$ / $0.677$ & $\textbf{0.457}$ / $0.456$\\
\midrule
7B & frozen model, generation alone & $0.826$ & $0.622$ & $0.625$ & $0.508$\\
7B & probe on: YOPO / two-pass & $\textbf{0.835}$ / $0.832$ & $\textbf{0.849}$ / $0.850$ & $\textbf{0.721}$ / $0.696$ & $\textbf{0.504}$ / $0.509$\\
7B & LoFiT on: YOPO / two-pass & $\textbf{0.837}$ / $0.836$ & $\textbf{0.869}$ / $0.863$ & $\textbf{0.711}$ / $0.687$ & $\textbf{0.493}$ / $0.512$\\
7B & LoReFT on: YOPO / two-pass & $\textbf{0.830}$ / $0.833$ & $\textbf{0.857}$ / $0.856$ & $\textbf{0.716}$ / $0.691$ & $\textbf{0.524}$ / $0.520$\\
\bottomrule
\end{tabular}
}
\end{table}

Three things hold across the \NCELLS{} (scale, writer, dataset) settings, which we call cells.
First, the writers raise free-form answering: answerable accuracy rises in every one of the
\NCELLS{} cells, by $0.7$ to $44$ points, and the probe's gains ($+0.7$ to $+17$ on 1.5B, $+7$ to
$+44$ on 3B, $+5$ to $+44$ on 7B) are of the same size as LoFiT's and LoReFT's ($+1$ to $+19$ on
1.5B, $+7$ to $+43$ on 3B, $+5$ to $+44$ on 7B). Second, every writer's edit lowers the
direction's AUROC on the generation pass where that AUROC has room to fall; in all \NLOSS{} cells where the loss exceeds $0.02$, the
refit map recovers between $71\%$ and all of it. Third, YOPO beats generation alone in
\NBEAT{} of the \NCELLS{} cells and is within $0.03$ of the two-pass reference in \NPAR{}, above
it in \NABOVE{}. The mechanism is the same everywhere: a strong writer stops the model from
abstaining on its own (the writer with no abstention at all scores only $0.15$--$0.55$, App.~\ref{app:suite}, Table~\ref{tab:m51interf}), so abstention has to come
from the read, and the direction, scored on the same pass after the map, recovers most of it: YOPO is $23$ to $48$
points above the writer alone in every cell. The two cells more than $0.03$ below the reference
are LoFiT on MuSiQue-Full and LoReFT on MuSiQue-C at 3B ($0.419$ vs.\ $0.450$; $0.645$ vs.\
$0.677$). YOPO stays below generation alone only on MuSiQue-Full (all three writers at 3B, the
probe and LoFiT at 7B, by $0.005$ to $0.064$), where the frozen model earns its score by
abstaining on its own and the writer removes that habit.

\section{Related Work}
\label{sec:related}
\textbf{Activation steering.} Fixed contrastive directions added to the residual stream
\citep{turner2023,panickssery2024}, inference-time shifts along learned directions
\citep{li2023iti}, representation engineering \citep{zou2023repe}, task and function vectors
\citep{todd2024,ilharco2023}, low-rank subspace rewrites \citep{wu2024reft}, localized
per-head offset tuning \citep{yin2024lofit}, and conditional
gating of a fixed vector \citep{lee2024cast}. Our probe instead makes the steering direction
depend on the residual it reads.
\textbf{Answerability and sufficiency reading.} Models linearly encode (un)answerability \citep{kadavath2022,slobodkin2023}; \citet{lavi2025} show, on datasets with human answerability labels, that a direction read from the hidden state detects insufficiency the model does not voice. We find (\S\ref{sec:reader}) that judgments trained into the model absorb the source domain's signature and degrade under
shift, consistent with the supervised-classifier collapse in \citet{lavi2025}. Reading the direction on a pass that a writer has edited is what neither line of work considers, and is what this paper adds.
\textbf{Reconstruction as repair.} Our map is trained like a denoiser
\citep{vincent2008,lehtinen2018}: reconstruct the unperturbed state from the perturbed one,
with the corruption process (the steering write) supplying free training pairs. The objective carries no task labels, which is what preserves transfer.

\section{Discussion}
\label{sec:discussion}
\textbf{Practical recipe.} If two forward passes are affordable, take the answer from a pass
with the probe on and the abstention score from a pass with it off (the two-pass reference);
if only one pass is, run YOPO, whose map closes most of the gap to that reference at every scale.
\textbf{Why the label-free map transfers.} The map inverts a perturbation whose geometry comes from the probe, not the
dataset. Once labels enter, task signature enters, and transfer degrades in proportion to how
much the labels helped in-domain (the direction's transfer matrix is App.~\ref{app:crossmatrix};
label-trained alternatives, App.~\ref{app:abstention}).
\textbf{Limitations.} On the \anli{} benchmark insufficiency is created by deleting text, and the deletion leaves a surface trace that a direction can read as early as layer 11
(\S\ref{sec:nativelabel}), before any sufficiency judgment could form; the architectural claims are anchored on
the human-label and four-QA results. Most experiments use Qwen2.5; the ten-backbone study covers one benchmark only, at $\leq$8B. Most settings are single-seed with a consistent scale trend; repeated runs agree within $0.002$, but full error bars on the two large grids were not computed.

\section{Conclusion}
A frozen model's residual stream can host a reasoning writer and a sufficiency reader in the
same forward pass if one rule is respected: decide abstention by reading the hidden state, never by training it, and train the map only to undo what the writer changes. The reconstruction map does
this: the system answers and abstains in one pass within $0.02$ of the two-pass alternative, on
the Qwen2.5 grid and nine of ten backbones from six families. On four QA datasets with
free-form answers it beats generation alone in \NBEAT{} of \NCELLS{} cells, by up to $36$
points, and stays within $0.03$ of the two-pass reference in \NPAR{}. Because the map is refit from unlabelled pairs, the writer is replaceable:
LoFiT and LoReFT drop in under the same recipe.

\subsection*{Reproducibility statement}
All datasets are public benchmarks (\anli{}, HellaSwag, PIQA, BoolQ, SQuAD~2.0, RepLiQA, MuSiQue), used via their
canonical releases. App.~\ref{app:data} lists the datasets, splits, item counts and checkpoints;
App.~\ref{app:scoring} gives the answer-correctness rule; Tables~\ref{tab:writersets} and~\ref{tab:fusedsets}
list the backbone, training data and evaluation data of every comparison, Table~\ref{tab:layers} the injection
and read layers of every backbone, and Table~\ref{tab:knobs} every fixed setting of YOPO. The reader protocol is
in \S\ref{sec:reader} and App.~\ref{app:readerscales}. \ifarxiv All code is available at \fi
\url{https://anonymous.4open.science/r/yopo-4FD2}.

\bibliography{paper}
\bibliographystyle{iclr2027_conference}

\appendix
\section{Ten Backbones, Six Families: Where the Repair Works}
\label{app:family}
The identical pipeline (probe, one-pass extraction, direction, map, benchmark) is rerun on
seven further backbones from five other families with nothing retuned: layers mapped by
proportional depth, Qwen hyper-parameters verbatim, one seed per cell
(Table~\ref{tab:family}): Qwen2.5 \citep{qwen2025qwen25}, OLMo-2 \citep{olmo2025olmo2},
TinyLlama \citep{zhang2024tinyllama}, StableLM-2 \citep{bellagente2024stablelm2}, SmolLM2
\citep{allal2025smollm2}, Phi-3-mini \citep{abdin2024phi3} and Mistral-7B
\citep{jiang2023mistral}. Six families by architecture: TinyLlama and SmolLM2 are both
Llama-architecture. The OLMo-2 and TinyLlama probes were re-trained for release (the originals
were not retained); their rows are the re-trained values.

\begin{table}[h]\centering\footnotesize
\setlength{\tabcolsep}{3.5pt}
\caption{Ten backbones (0.5B--7B), \anli{} four-condition validation, three-way accuracy at
$p{=}25$ (left) and the direction's AUROC (full vs.\ replace-both) on the probe-off pass, the
probe-on pass and the map-repaired pass (right). \textbf{Bold}: YOPO $\geq$ the two-pass reference. Single seed per
cell.}
\label{tab:family}
\resizebox{\textwidth}{!}{%
\begin{tabular}{ll cccc ccc}
\toprule
backbone & size & frozen model & probe alone & two-pass ref.\ & \textbf{YOPO (1 pass)} & \multicolumn{3}{c}{direction AUROC: probe off / on / after map}\\
\midrule
Qwen2.5 & 0.5B & 0.318 & 0.420 & 0.545 & \textbf{0.548} & 0.905 & 0.828 & 0.873\\
OLMo-2 & 1B & 0.485 & 0.505 & 0.635 & \textbf{0.638} & 0.899 & 0.894 & 0.903\\
TinyLlama (Llama arch.) & 1.1B & 0.375 & 0.375 & 0.500 & \textbf{0.500} & 0.914 & 0.812 & 0.908\\
Qwen2.5 & 1.5B & 0.375 & 0.590 & 0.753 & \textbf{0.753} & 0.962 & 0.913 & 0.954\\
StableLM-2 & 1.6B & 0.375 & 0.480 & 0.625 & 0.618 & 0.931 & 0.904 & 0.948\\
SmolLM2 (Llama arch.) & 1.7B & 0.375 & 0.580 & 0.778 & 0.755 & 0.958 & 0.823 & 0.914\\
Qwen2.5 & 3B & 0.563 & 0.625 & 0.790 & \textbf{0.800} & 0.967 & 0.938 & 0.968\\
Phi-3-mini & 3.8B & 0.375 & 0.375 & 0.578 & 0.575 & 0.962 & 0.939 & 0.963\\
Qwen2.5 & 7B & 0.628 & 0.668 & 0.863 & 0.858 & 0.985 & 0.995 & 0.986\\
Mistral & 7B & 0.375 & 0.375 & 0.528 & \textbf{0.530} & 0.940 & 0.933 & 0.941\\
\bottomrule
\end{tabular}
}
\end{table}

The map restores the direction on all ten backbones. AUROC here is full vs.\
replace-both on the validation split, the definition of Table~\ref{tab:writers}, from a
separate extraction of the same states (cells differ from Table~\ref{tab:writers} by at most
$0.011$). Probes here use the 5k-item, 1-epoch budget, so
Phi-3-mini's probe adds no answer accuracy, whereas the 15k-item, 3-epoch probe of
\S\ref{sec:writer} gains $2.9$ points. The direction's AUROC is restored on every backbone, from
$0.828\to0.873$ on Qwen-0.5B, $0.812\to0.908$ on TinyLlama, $0.823\to0.914$ on SmolLM2 and
$0.939\to0.963$ on Phi-3-mini to within $0.01$ of the probe-off pass on Qwen 1.5B and 3B; on
Qwen-7B the edit had raised the read ($0.985\to0.995$) and the map returns it to $0.986$.
Under the pre-registered random start the optimisation did not move at all on SmolLM2 and
Phi-3-mini (App.~\ref{app:failures}, item (x)). End to end, YOPO is within $0.005$ three-way
accuracy of the two-pass reference on eight of the ten backbones, spanning five families
(above it on Qwen-0.5B, OLMo-2, Qwen-3B and Mistral), $0.007$ below on StableLM-2 and
$0.023$ below on SmolLM2, so within $0.02$ on nine of the ten, spanning all six families. The frozen baseline is
a degenerate $0.375$ on six of the ten, and on three
of them (TinyLlama, Phi-3-mini, Mistral) the probe adds no answer accuracy at all, so what the
system adds there is the threshold alone ($+12$ to $+20$ points: TinyLlama $0.375\to0.500$, Phi-3-mini
$0.375\to0.575$, Mistral $0.375\to0.530$). The two stalls under the random start are an
optimisation defect of that recipe, not a capacity limit: App.~\ref{app:failures}, item (x),
records them, and the identity start that repairs them is the recipe of every table.

\section{The Steering Injection-Geometry Grid}
\label{app:steergrid}

\textbf{Ablations behind \S\ref{sec:writer}.} In a position ablation on 1.5B \anli{} (App.~\ref{app:data},
settings table), reading five spread positions beats the last token alone by $5.6$ accuracy
points at three layers and $\alpha_{\max}{=}0.3$ ($65.5$ vs.\ $71.1$, Table~\ref{tab:steergrid}); in a
separate run of that configuration, two, three and four positions keep $71\%$, $94\%$ and $90\%$
of the five-position gain over the last token ($66.0$ vs.\ $70.5$); the number of probe layers is a weak
lever (App.~\ref{app:steergrid}); replacing the learned magnitude by a constant changes
accuracy by $0.26$ points ($p{=}0.767$); and training data saturates early: 5k items, 1 epoch
gives $+19.7$ and 100k items, 3 epochs $+22.9$.

This grid sweeps the probe's configuration (layer count, injection positions, magnitude cap)
across three datasets and the scale ladder. Spread positions beat last-token injection at every
layer count on \anli{} (1.5B \anli{} $+20.2$ vs.\ $+14.7$ at three layers), layer count is a weak lever,
and gains are largest where the frozen baseline is weakest. Datasets: \anli{}, HellaSwag and
PIQA. Each probe is trained on 5k items of its task (seed 0) and evaluated on 1{,}532 held-out
items against the frozen baseline \bz{}. ``--'' cells are configurations outside the
registered sweep, not pending runs.

\begin{table}[h]\centering
\small
\setlength{\tabcolsep}{4pt}
\caption{Steering grid: accuracy (\%$\Delta$ vs.\ the frozen baseline \bz{}) across injection
layer count, position scheme, and $\alpha_{\max}$, per dataset and backbone.}
\label{tab:steergrid}
\begin{tabular}{llccc}
\toprule
config & $\alpha_{\max}$ & \anli{} & HellaSwag & PIQA\\
\midrule
\textbf{Qwen2.5-1.5B} frozen \bz{} & -- & $50.8$ & $74.1$ & $68.6$\\[1pt]
\quad 1 layer, last & $0.3$ & $65.6$ {\scriptsize($+14.8$)} & $74.9$ {\scriptsize($+0.8$)} & $70.6$ {\scriptsize($+2.0$)}\\
\quad 1 layer, last & $0.5$ & $65.5$ {\scriptsize($+14.6$)} & $74.5$ {\scriptsize($+0.4$)} & $72.6$ {\scriptsize($+4.0$)}\\
\quad 1 layer, spread-5 & $0.3$ & $69.1$ {\scriptsize($+18.3$)} & $75.8$ {\scriptsize($+1.7$)} & $73.4$ {\scriptsize($+4.8$)}\\
\quad 1 layer, spread-5 & $0.5$ & $70.4$ {\scriptsize($+19.5$)} & $76.2$ {\scriptsize($+2.1$)} & $75.8$ {\scriptsize($+7.2$)}\\
\quad 3 layers, last & $0.3$ & $65.5$ {\scriptsize($+14.7$)} & $74.7$ {\scriptsize($+0.6$)} & $72.7$ {\scriptsize($+4.1$)}\\
\quad 3 layers, last & $0.5$ & $65.9$ {\scriptsize($+15.0$)} & $74.5$ {\scriptsize($+0.4$)} & $72.5$ {\scriptsize($+3.9$)}\\
\quad 3 layers, spread-5 & $0.3$ & $71.1$ {\scriptsize($+20.2$)} & $77.3$ {\scriptsize($+3.2$)} & $74.7$ {\scriptsize($+6.1$)}\\
\quad 3 layers, spread-5 & $0.5$ & $72.5$ {\scriptsize($+21.6$)} & $77.7$ {\scriptsize($+3.6$)} & $76.0$ {\scriptsize($+7.4$)}\\
\quad 5 layers, last & $0.3$ & $65.5$ {\scriptsize($+14.6$)} & $75.1$ {\scriptsize($+1.0$)} & $71.9$ {\scriptsize($+3.3$)}\\
\quad 5 layers, last & $0.5$ & $66.1$ {\scriptsize($+15.2$)} & -- & --\\
\quad 5 layers, spread-5 & $0.3$ & $70.4$ {\scriptsize($+19.6$)} & $77.3$ {\scriptsize($+3.1$)} & $76.5$ {\scriptsize($+7.9$)}\\
\quad 5 layers, spread-5 & $0.5$ & $71.2$ {\scriptsize($+20.4$)} & -- & --\\
\quad 7 layers, last & $0.3$ & $66.2$ {\scriptsize($+15.3$)} & $75.5$ {\scriptsize($+1.3$)} & $72.5$ {\scriptsize($+3.9$)}\\
\quad 7 layers, last & $0.5$ & $66.6$ {\scriptsize($+15.8$)} & $75.3$ {\scriptsize($+1.1$)} & $72.2$ {\scriptsize($+3.6$)}\\
\quad 7 layers, spread-5 & $0.3$ & $71.3$ {\scriptsize($+20.4$)} & $77.5$ {\scriptsize($+3.3$)} & $75.9$ {\scriptsize($+7.3$)}\\
\quad 7 layers, spread-5 & $0.5$ & $72.4$ {\scriptsize($+21.5$)} & $78.5$ {\scriptsize($+4.3$)} & $74.8$ {\scriptsize($+6.2$)}\\
\midrule
\textbf{Qwen2.5-3B} frozen \bz{} & -- & $66.6$ & $63.0$ & $74.9$\\[1pt]
\quad 1 layer, last & $0.3$ & $72.5$ {\scriptsize($+5.9$)} & $81.7$ {\scriptsize($+18.7$)} & $77.9$ {\scriptsize($+3.0$)}\\
\quad 1 layer, last & $0.5$ & $72.4$ {\scriptsize($+5.7$)} & $81.9$ {\scriptsize($+18.9$)} & $77.5$ {\scriptsize($+2.6$)}\\
\quad 1 layer, spread-5 & $0.3$ & $74.5$ {\scriptsize($+7.9$)} & $82.9$ {\scriptsize($+19.9$)} & $80.9$ {\scriptsize($+5.9$)}\\
\quad 1 layer, spread-5 & $0.5$ & $74.9$ {\scriptsize($+8.3$)} & $84.1$ {\scriptsize($+21.1$)} & $81.3$ {\scriptsize($+6.4$)}\\
\quad 3 layers, last & $0.3$ & $72.8$ {\scriptsize($+6.1$)} & $82.5$ {\scriptsize($+19.5$)} & $78.3$ {\scriptsize($+3.4$)}\\
\quad 3 layers, last & $0.5$ & $72.7$ {\scriptsize($+6.1$)} & $84.0$ {\scriptsize($+21.0$)} & $78.9$ {\scriptsize($+4.0$)}\\
\quad 3 layers, spread-5 & $0.3$ & $76.2$ {\scriptsize($+9.5$)} & $84.0$ {\scriptsize($+21.0$)} & $80.6$ {\scriptsize($+5.7$)}\\
\quad 3 layers, spread-5 & $0.5$ & $75.5$ {\scriptsize($+8.8$)} & $84.1$ {\scriptsize($+21.1$)} & $80.8$ {\scriptsize($+5.9$)}\\
\quad 5 layers, last & $0.3$ & $73.3$ {\scriptsize($+6.7$)} & $82.0$ {\scriptsize($+19.0$)} & $78.5$ {\scriptsize($+3.5$)}\\
\quad 5 layers, last & $0.5$ & $73.7$ {\scriptsize($+7.0$)} & -- & --\\
\quad 5 layers, spread-5 & $0.3$ & $75.1$ {\scriptsize($+8.5$)} & $83.9$ {\scriptsize($+20.9$)} & $81.1$ {\scriptsize($+6.2$)}\\
\quad 5 layers, spread-5 & $0.5$ & $75.8$ {\scriptsize($+9.2$)} & -- & --\\
\quad 7 layers, last & $0.3$ & $74.2$ {\scriptsize($+7.5$)} & $81.7$ {\scriptsize($+18.7$)} & $77.5$ {\scriptsize($+2.6$)}\\
\quad 7 layers, last & $0.5$ & $73.6$ {\scriptsize($+6.9$)} & $84.5$ {\scriptsize($+21.5$)} & $78.5$ {\scriptsize($+3.5$)}\\
\quad 7 layers, spread-5 & $0.3$ & $74.9$ {\scriptsize($+8.3$)} & $84.5$ {\scriptsize($+21.5$)} & $80.3$ {\scriptsize($+5.4$)}\\
\quad 7 layers, spread-5 & $0.5$ & $76.4$ {\scriptsize($+9.7$)} & $82.6$ {\scriptsize($+19.6$)} & $80.1$ {\scriptsize($+5.2$)}\\
\midrule
\textbf{Qwen2.5-7B} frozen \bz{} & -- & $77.6$ & $86.0$ & $84.7$\\[1pt]
\quad 3 layers, last & $0.3$ & $77.9$ {\scriptsize($+0.3$)} & $85.9$ {\scriptsize($-0.1$)} & $85.0$ {\scriptsize($+0.3$)}\\
\quad 3 layers, last & $0.5$ & $77.9$ {\scriptsize($+0.3$)} & -- & --\\
\quad 3 layers, spread-5 & $0.3$ & $80.5$ {\scriptsize($+2.9$)} & $88.2$ {\scriptsize($+2.2$)} & $85.1$ {\scriptsize($+0.4$)}\\
\quad 3 layers, spread-5 & $0.5$ & $80.7$ {\scriptsize($+3.1$)} & -- & --\\
\quad 5 layers, last & $0.3$ & $78.5$ {\scriptsize($+0.8$)} & $86.0$ {\scriptsize($+0.0$)} & $85.4$ {\scriptsize($+0.7$)}\\
\quad 5 layers, last & $0.5$ & $78.7$ {\scriptsize($+1.1$)} & -- & --\\
\quad 5 layers, spread-5 & $0.3$ & $80.8$ {\scriptsize($+3.2$)} & $88.8$ {\scriptsize($+2.8$)} & $85.1$ {\scriptsize($+0.5$)}\\
\quad 5 layers, spread-5 & $0.5$ & $81.1$ {\scriptsize($+3.5$)} & -- & --\\
\quad 7 layers, last & $0.3$ & $78.4$ {\scriptsize($+0.8$)} & $86.1$ {\scriptsize($+0.1$)} & $85.4$ {\scriptsize($+0.7$)}\\
\quad 7 layers, last & $0.5$ & $79.2$ {\scriptsize($+1.6$)} & -- & --\\
\quad 7 layers, spread-5 & $0.3$ & $80.6$ {\scriptsize($+3.0$)} & $88.2$ {\scriptsize($+2.2$)} & $85.8$ {\scriptsize($+1.2$)}\\
\quad 7 layers, spread-5 & $0.5$ & $81.0$ {\scriptsize($+3.4$)} & -- & --\\
\bottomrule
\end{tabular}
\end{table}

\section{LoFiT and LoReFT at Equal Budget}
\label{app:lofit}
\textbf{Writer-transfer round.} In a separate pre-registered
writer-transfer round with our reimplementation of LoFiT and the official LoReFT code (HellaSwag/BoolQ,
answers read from the two option tokens' logits, App.~\ref{app:lofit}), the probe is the only writer that gains
over frozen on both datasets at 7B, while LoFiT loses $2.6$--$2.7\pp$ and LoReFT collapses
everywhere off-domain; at 1.5B HellaSwag reverses ($+0.7$ for LoFiT, $-3.6$ for the probe), so
the verdict is mixed (three of four contrasts). Whether a write transfers
is not decided by the space it writes in: the three activation-space writers span $22$ points
on HellaSwag at 7B.

\begin{table}[h]\centering\footnotesize
\setlength{\tabcolsep}{3pt}
\caption{Matched-budget comparison on Qwen2.5-1.5B, all trained on \anli{} (15k items, 3
epochs, seed 0). The ReFT-style row is our reimplementation of a rank-128
subspace edit under this protocol; the official LoReFT is compared in App.~\ref{app:lofit}.}
\label{tab:matched}
\begin{tabular}{llccc}
\toprule
system & space & params & \anli{} & HellaSwag (zero-shot)\\
\midrule
frozen backbone & -- & 0 & $50.78$ & $73.60$\\
steering probe (ours) & activation & $3.58$M & $72.52$ ($+21.74$) & $69.87$ ($-3.73$)\\
ReFT-style $r{=}128$ (ours) & activation & $1.18$M & $73.43$ ($+22.65$) & $71.73$ ($-1.87$)\\
LoRA $r{=}3$ & weight & $3.46$M & $78.46$ ($+27.68$) & $79.53$ ($+5.93$)\\
\bottomrule
\end{tabular}
\end{table}

The nearest neighbours of the probe are localised representation-editing fine-tuners of frozen
backbones. LoFiT \citep{yin2024lofit} selects a sparse subset of attention heads and learns
\emph{static} per-head offset vectors; LoReFT \citep{wu2024reft} learns low-rank subspace
rewrites of the residual stream at fixed positions. The probe differs in making the edit an
\emph{input-conditioned function} of the residual it reads. Table~\ref{tab:lofit} is a
pre-registered comparison on \anli{}: under the shared frozen-backbone protocol both baselines
are strong, and the probe's advantage does not survive at matched data budgets. Neither budget
control changes the verdict: three epochs \emph{lower} the probe at 1.5B/7B, so the baselines'
advantage is not explained by their extra epochs, and at matched total data cost LoFiT still
leads the probe by $1.7\pp$ at 1.5B.

\begin{table}[h]\centering
\small
\setlength{\tabcolsep}{5pt}
\caption{Answer accuracy (\%) on \anli{} held-out validation (1{,}532 items) across the
Qwen2.5 scale ladder, all methods trained on the same 5k items (seed 0) with the frozen
backbone and evaluated by the same two-token protocol. Probe rows are the injection-geometry
grid cells (Table~\ref{tab:steergrid}, 1 epoch; the 3-epoch row repeats the identical
configuration with the baselines' epoch budget). The matched-cost LoFiT row (1.5B) compresses
selection to 1{,}250 items and bias-tuning to one epoch on 3{,}750, so its total gradient data
equals the probe's single 5k-item pass. LoFiT: faithful reimplementation per the paper
(official code targets Llama/Gemma), top-10\% attention-head offset vectors, 4.4k/7.4k/10.0k
trainable parameters at 1.5B/3B/7B, 3 epochs. LoReFT: official pyreft intervention, commonsense
recipe (all layers, $r{=}8$, f7+l7, shared), 0.69M/1.18M/1.61M parameters, 3 epochs.
\textsuperscript{$\dagger$}At 3B the official LoReFT recipe (lr $9{\times}10^{-4}$, tuned for
Llama-7B) fails to converge (training loss plateaus at $\ln 2$); a single pre-declared
sensitivity rerun at lr $3{\times}10^{-4}$ reaches $79.7$. Both numbers are reported.}
\label{tab:lofit}
\begin{tabular}{lccc}
\toprule
method & \anli{} 1.5B & \anli{} 3B & \anli{} 7B\\
\midrule
frozen baseline \bz{} & $50.8$ & $66.6$ & $77.6$\\
probe (3 layers, spread-5, $\alpha_{\max}{=}0.5$; 1 ep) & $72.5$ & $75.5$ & $80.7$\\
probe (same config; 3 ep, equal budget) & $70.7$ & $76.0$ & $80.3$\\
LoFiT (reimpl.) \citep{yin2024lofit} & $72.6$ & $78.3$ & $81.4$\\
LoFiT (matched cost: one total data pass, as probe@1 ep) & $74.2$ & -- & --\\
LoReFT (official) \citep{wu2024reft} & $76.0$ & $50.9$\textsuperscript{$\dagger$} & $82.2$\\
\bottomrule
\end{tabular}
\end{table}

\textbf{Off-domain transfer of the writers.} Table~\ref{tab:lofittransfer} evaluates the
\anli{}-trained writers zero-shot on HellaSwag and BoolQ against each scale's frozen baseline.
The verdict on the pre-registered hypothesis (conditional writes transfer better than static
ones) is \emph{mixed}. At 7B the probe is the only writer that gains over frozen on both
datasets ($+0.9\pp$ HellaSwag, $+0.6\pp$ BoolQ) while LoFiT loses $2.6$--$2.7\pp$; at 1.5B BoolQ
agrees but HellaSwag reverses ($+0.7\pp$ for LoFiT vs.\ $-3.6\pp$ for the probe). LoReFT, the
strongest in-domain writer, collapses off-domain everywhere ($-21.5\pp$ HellaSwag at 7B;
$-12.0/-16.3\pp$ BoolQ), echoing the ladder of \S\ref{sec:reader}: the more freely a writer
rewrites the stream, the more task signature it absorbs and the worse it travels.

\begin{table}[h]\centering
\small
\setlength{\tabcolsep}{5pt}
\caption{Zero-shot transfer of \anli{}-trained writers. Answer accuracy (\%) on the HellaSwag
binary set (1{,}500 items) and BoolQ (1{,}499 items), Qwen2.5-1.5B/7B, same two-token
protocol; writers are those of Table~\ref{tab:lofit}, unchanged (probe = the 1-epoch grid
cell), eval-only. Pre-registered as H5; verdict mixed (three of four scale$\times$dataset
contrasts favour the conditional write; 1.5B HellaSwag reverses).}
\label{tab:lofittransfer}
\begin{tabular}{lcccc}
\toprule
& \multicolumn{2}{c}{HellaSwag} & \multicolumn{2}{c}{BoolQ}\\
\cmidrule(lr){2-3}\cmidrule(lr){4-5}
writer (trained on \anli{}) & 1.5B & 7B & 1.5B & 7B\\
\midrule
frozen baseline \bz{} & $74.1$ & $86.0$ & $72.2$ & $81.7$\\
probe & $70.5$ & $86.9$ & $69.4$ & $82.3$\\
LoFiT & $74.8$ & $83.4$ & $66.5$ & $79.1$\\
LoReFT & $70.0$ & $64.5$ & $60.2$ & $65.4$\\
\bottomrule
\end{tabular}
\end{table}

All cells, hypothesis verdicts (including the failed equal-budget and matched-cost
pre-registrations), deviations, and wall-clocks are in the pre-registration document in the
\ifarxiv code repository\fi.

\section{The Direction-Transfer Cross-Matrix}
\label{app:crossmatrix}
Table~\ref{tab:crossmatrix} fits the sufficiency direction on each of five datasets and scores
it on every other one, computed entirely from our stored probe-off states. To \anli{} and HellaSwag (\S\ref{sec:yopo}) and SQuAD2 and RepLiQA
(\S\ref{sec:nativelabel}) we add a MuSiQue set with native answerability labels (2{,}000 training and 1{,}000 evaluation items, balanced; every context
window starts with the supporting paragraphs, so the gold reasoning chain is never cut off).
The table's AUROC uses a different label and a different layer rule than
Table~\ref{tab:writers} (see the caption), so its cells are not comparable to the main text's.

At 7B, directions fit on native labels rank our constructed minimal pairs and vice versa
(HellaSwag$\to$SQuAD2 $0.930$; SQuAD2$\to$\anli{} $0.801$; RepLiQA$\to$HellaSwag $0.914$).
At 1.5B the single-direction picture degrades, and the same trend shows in the raw vectors:
the cosine between the \anli{}-fit and the HellaSwag-fit direction at layer 18 rises from
$0.58$ on 1.5B to $0.75$ on 7B. MuSiQue splits by role: as a target it is the hardest column
(no source exceeds $0.830$; in domain only $0.850$), yet at 7B it is the strongest source
row (all four transfer cells $\geq 0.916$). In decision accuracy (the fraction of items whose
sufficient/insufficient call is correct, threshold at the label-free median of the target's
own scores; Qwen2.5-7B), the direction fit on MuSiQue and scored on the probe-off pass
transfers to RepLiQA at $0.917$ and to SQuAD2 at $0.844$; scored on the probe-on pass after
the map it gives $0.885$ and $0.850$; the reverse direction, fit on SQuAD2 and scored on
MuSiQue, reaches $0.742$.

\textbf{A judgment trained on labels.} The label-trained judgment of \S\ref{sec:reader} in its
LoRA form (a rank-3 LoRA adapter on Qwen2.5-7B-Instruct, fine-tuned on labelled items to answer
whether the context suffices) loses decision accuracy off domain: trained on SQuAD2 it scores
$0.892$ on the 2{,}000 SQuAD2 test items in domain and $0.756$ on the 2{,}000 RepLiQA test items
(Table~\ref{tab:reader7b}), a $13.7\pp$ drop.

\begin{table}[t]
\centering
\caption{\textbf{Direction-transfer cross-matrix: AUROC of the sufficiency direction,
Qwen2.5-1.5B and 7B-Instruct, computed from our stored probe-off states.} Row: the dataset
the direction is fit on; column: the dataset it is scored on; diagonal (gray) $=$ in domain.
Fitting: on a random 80\% of the dataset's training rows, the difference of class means at
each of six candidate layers $\{11,12,14,16,18,20\}$ (the layers stored for all five
datasets); the layer with the highest AUROC on the remaining 20\% is kept and frozen with
the direction. Scoring: AUROC of that frozen direction, at that layer, on the target's
validation rows (\anli{} 400, HellaSwag 500, MuSiQue 1{,}000; SQuAD2 and RepLiQA their
judgment-prompt validation rows), probe off, no recalibration. Labels: on \anli{} and
HellaSwag (our 4-condition sets) sufficient $=$ the three answerable versions against the
replaced one; SQuAD2 and RepLiQA carry native answerable/unanswerable labels; MuSiQue
(multi-hop, native labels, context windows starting with the supporting paragraphs). This
construction differs from Table~\ref{tab:writers} in the label (there: full vs.\
replace-both only) and in the layer (there: the fixed read layer 18 on 1.5B), so the cells
are not comparable to it: \anli{}$\to$HellaSwag at 1.5B is $.748$ here and $0.919$ there,
and the \anli{} diagonal is $.986$ here and $0.962$ there. One deterministic fit per cell,
no $\pm$. Bottom row: mean and minimum over the 20 off-diagonal cells of each model.}
\label{tab:crossmatrix}
\small
\setlength{\tabcolsep}{3.4pt}
\resizebox{\textwidth}{!}{%
\begin{tabular}{l ccccc c ccccc}
\toprule
& \multicolumn{5}{c}{\textbf{Qwen2.5-1.5B} (scored on)} & &
  \multicolumn{5}{c}{\textbf{Qwen2.5-7B} (scored on)} \\
\cmidrule{2-6}\cmidrule{8-12}
fit on & $\alpha$NLI & Hella & SQuAD2 & RepLiQA & MuSiQue & & $\alpha$NLI & Hella & SQuAD2 & RepLiQA & MuSiQue \\
\midrule
$\alpha$NLI & \textcolor{gray}{.986} & .748 & .648 & .465 & .677 & & \textcolor{gray}{.995} & .928 & .805 & .795 & .766 \\
HellaSwag   & .923 & \textcolor{gray}{.905} & .648 & .480 & .657 & & .967 & \textcolor{gray}{.959} & .930 & .905 & .808 \\
SQuAD2      & .766 & .872 & \textcolor{gray}{.863} & .941 & .817 & & .801 & .828 & \textcolor{gray}{.932} & .961 & .815 \\
RepLiQA     & .919 & .599 & .552 & \textcolor{gray}{.999} & .616 & & .867 & .914 & .913 & \textcolor{gray}{.988} & .830 \\
MuSiQue     & .799 & .864 & .865 & .949 & \textcolor{gray}{.832} & & .918 & .935 & .916 & .969 & \textcolor{gray}{.850} \\
\midrule
\multicolumn{6}{l}{off-diagonal mean\,/\,min: \ .740\,/\,.465} & &
\multicolumn{5}{l}{.879\,/\,.766} \\
\bottomrule
\end{tabular}}
\end{table}

\section{Training the Abstention In: Label-Trained Variants, Not Part of YOPO}
\label{app:abstention}
This appendix reports variants that train the abstention on labels. None of them is part of
YOPO: a probe trained with a third, ``cannot answer'' output (the \emph{abstention-trained
probe}), and a \emph{supervised judge}, a small MLP trained with binary cross-entropy (BCE)
on sufficiency labels to score a hidden state. They are kept as negative results and as
comparison points for \S\ref{sec:nativelabel}. Conventions: the backbone is
Qwen2.5-7B-Instruct unless a size is named; ``three-way accuracy'' is the metric of
\S\ref{sec:benchmarks} (answer 1, answer 2, abstain), scored on the four-condition validation
sets (\anli{} 400 rows; HellaSwag 500 rows, nothing fit on it) and on the MuSiQue evaluation
set of App.~\ref{app:crossmatrix} (1{,}000 rows); ``in domain'' means the probe's own
training set, ``transfer'' any other. ``Abstention recall'' is the fraction of insufficient
rows abstained on, ``over-abstention'' the fraction of answerable rows abstained on, and
``gate AUROC'' the AUROC of an abstention score for insufficient vs.\ answerable rows, all on
the same rows. ``Probe-on'' and ``probe-off'' name the pass that is scored. A
\emph{calibrated} score is one thresholded at the label-free 25th percentile of the evaluated
set's own scores, as YOPO does, instead of by argmax; the \emph{calibrated prompt baseline}
is the frozen model, probe off, under a three-option prompt (answer 1, answer 2, cannot
answer), with its probability of ``cannot answer'' calibrated the same way. Every $\pm$ is
the sample standard deviation over three probe seeds, except the two supervised-judge values
$0.793\pm0.003$ and $0.878\pm0.007$, which are population standard deviations over three
seeds ($0.004$ and $0.008$ as sample standard deviations).

\textbf{Setup and headline results.} The abstention-trained probe has a three-way head, the
same architecture and training budget as the answer-only probe, and is trained once on
\anli{} and once on MuSiQue. At 1.5B it fails both ways: the \anli{}-trained probe on
HellaSwag abstains on only $0.08$ of the insufficient rows (abstention recall), and the
MuSiQue-trained probe over-abstains on $91\%$ of \anli{}'s and $70\%$ of HellaSwag's
answerable rows. At 7B both failures vanish: three-way accuracy in domain is $0.880$ on
\anli{} (above the two-pass reference $0.863$ and YOPO's $0.858$ of Table~\ref{tab:combined}) and $0.872$ on MuSiQue,
with over-abstention $0.017$ and $0.10$; \anli{}$\to$HellaSwag reaches $0.726$ with
abstention recall $0.848$. Trained abstention does transfer at 7B, but transferring is not
winning: MuSiQue$\to$HellaSwag lands at $0.682$, below the frozen model's own three-option
prompt by argmax ($0.702$).

\textbf{Where the interference lives.} Under transfer (\anli{}-trained probe scored on
HellaSwag, 7B) the trained abstention score ranks the rows equally well on the probe-on and
on the probe-off pass (gate AUROC $0.9337$ vs.\ $0.9338$): the transfer loss sits on the
answer side and in the decision rule, not in the abstention score.

\textbf{Decomposing the transfer failure} (post hoc on stored states, then pre-registered
follow-ups; 7B). The abstention-trained probe commits to ``cannot answer'' by argmax, an
uncalibrated threshold frozen into its weights. Re-reading its own probability of ``cannot
answer'' as a score and calibrating it recovers $6$--$8\pp$ of three-way accuracy everywhere
(MuSiQue$\to$HellaSwag $0.682\to0.744$; MuSiQue$\to$\anli{} $0.7325\to0.810$; gate AUROC of
the trained score $0.918$--$0.970$): most of the transfer gap was a decision-rule failure, not
a representation failure. The remainder separates the write's value by domain. In domain,
the calibrated probe-on pass beats the calibrated probe-off pass by $7$--$8\pp$ ($0.864$ vs.\
$0.790$ on MuSiQue; $0.885$ vs.\ $0.8125$ on \anli{}; these calibrated in-domain values
differ from the argmax values $0.872$ and $0.880$ above), so training does refine the
in-domain partition. On transfer its net contribution is zero to negative: $0.744$ against
the calibrated prompt baseline's $0.784$ on HellaSwag (MuSiQue-trained), and $0.810$ against
$0.815$ on \anli{}. The calibrated prompt baseline is itself a strong baseline we have not
seen reported; its prompt differs from the two-option judgment prompts of
\S\ref{sec:nativelabel}, so the two are not comparable across constructions.

\textbf{Two pre-registered repair candidates} (7B, MuSiQue-trained; three-way accuracy on
HellaSwag unless stated). \emph{D1, a raw read--write split}: the write stays answer-only and
the abstention becomes a calibrated read of a direction fit on the probe-on states. It fails
both of its pre-registered lines ($0.624$ against targets $>0.744$ and $>0.784$), and for the
reason \S\ref{sec:problem} predicts: without the map, the probe's edit swamps the transfer
read (gate AUROC $0.917$ on the probe-off pass $\to0.858$ on the probe-on pass; accuracy on
answerable rows $0.820\to0.719$). The split itself is sound (read on the probe-off pass it
scores $0.774$); reading the probe-on pass without correction is exactly the write--read
interference of \S\ref{sec:problem}. \emph{D2, a two-source probe} (\anli{}$+$MuSiQue 50/50
at the same 2{,}000-item budget) repairs what it targets and no more: over-abstention falls
$0.269\to0.099$ on HellaSwag and $0.26\to0.047$ on \anli{} (abstention recall $0.96$); in
domain it stays near the single-source values with half the per-source data ($0.870$ on
\anli{}, $0.845$ on MuSiQue); and its argmax equals its calibrated score on HellaSwag
($0.744{=}0.744$), so source diversity internalises the threshold that single-source training
leaves miscalibrated. But it exactly ties its pre-registered transfer line ($0.744$), still
short of the calibrated prompt baseline ($0.784$).

\textbf{Four write-side repairs} (7B, \anli{}-trained probe, scored on HellaSwag, probe on).
Down-weighting the abstain class in the loss ($\lambda{=}0.25/0.5$: accuracy on answerable
rows $0.789/0.781$, gate AUROC $0.900/0.920$), a KL anchor that penalises drift of the answer
distribution (best accuracy on answerable rows, $0.811$; gate AUROC $0.897$), and PCGrad
gradient surgery between the two losses (accuracy on answerable rows $0.787$; gate AUROC
$0.891$; the worst calibrated three-way accuracy, $0.716$) all remove the write--read
interference and all pay in gate quality: within one shared write vector, answer fidelity
and abstention quality trade at par under shift. The best write-side calibrated three-way
accuracy is $\lambda{=}0.5$'s $0.746$, $+1.0\pp$ over the unrepaired probe's $0.736$. The
repair that avoids the trade, moving the abstention out of the write entirely, is the
supervised judge below.

\textbf{The read-side repair: a supervised judge.} Keep the write answer-only and train a
small MLP ($3584{\to}128{\to}1$, 459K parameters) with BCE on the four-condition
answerable-vs-not label to score the layer-18 residual of the probe-on pass, with no
gradient into the write. On HellaSwag (7B, \anli{}-trained) its gate AUROC is $0.955$ on the
probe-on pass and $0.964$ on the probe-off pass, against $0.934$ for the abstention-trained
probe's own score on the same rows. End to end it gives the best transfer three-way accuracy
we measure, $0.793\pm0.003$ on HellaSwag, but on the probe-off pass, which costs a second
forward pass (the calibrated prompt baseline: $0.784$); in domain on \anli{}, scored on the
probe-on pass, it reaches $0.878\pm0.007$, above the two-pass reference $0.863$. The judge
is scale-dependent: its gate AUROC on HellaSwag on the probe-on pass is $0.627$ at 1.5B,
$0.825$ at 3B and $0.955$ at 7B.

\textbf{The supervised-judge variant: write and read on separate channels, one pass.}
Teaching the write to suppress itself on unlabelled out-of-distribution prompts gives the
configuration of Figure~\ref{fig:yopo2}: an answer-only write whose learned magnitude is
trained to go to zero on unlabelled out-of-distribution prompts (a \emph{self-muting}
magnitude; $\lambda$ weights the suppression term), the supervised judge for abstention, a
label-free threshold, one forward pass. At $\lambda{=}0.3$ (three seeds, 7B, \anli{}-trained,
probe on) it scores $0.877\pm0.008$ three-way accuracy in domain on \anli{}, above the
two-pass reference $0.863$ on every seed, and $0.781\pm0.012$ on HellaSwag: parity with the
calibrated prompt baseline ($0.784$). Stronger suppression ($\lambda{=}0.5/1.0$:
$0.782/0.786$ on HellaSwag, single seed) stays within seed noise of $\lambda{=}0.3$ and costs
in-domain accuracy ($0.860/0.8575$). Two further suppression objectives close the question:
a hinge that penalises only out-of-distribution magnitudes above a floor (three seeds: in
domain $0.859\pm0.018$, HellaSwag $0.781\pm0.013$) and a two-sided objective that also holds
in-domain magnitudes up (in domain $0.883\pm0.008$, above the two-pass reference on every
seed and the best in-domain configuration we have; HellaSwag $0.764\pm0.014$). Three
objectives trace one monotone trade: each increment of in-domain gain from the write costs a
comparable increment on HellaSwag, while the judge's three-way accuracy on the probe-off pass
stays configuration-independent at $0.793$--$0.797$. With the write active, beating the
calibrated prompt baseline under shift was out of reach for this architecture at 7B.
Together with the write-side repairs, this is the design rule the main text follows:
abstention belongs on the read side.

\begin{figure}[h]\centering
\begin{tikzpicture}[font=\scriptsize, >={Stealth[length=1.6mm]}, node distance=2.0mm,
  blk/.style={draw, rounded corners=2.5pt, minimum width=30mm, minimum height=4.8mm, align=center, line width=0.7pt},
  frz/.style={blk, fill=figgray, draw=figgrayln},
  inj/.style={blk, fill=figblue, draw=figblueln, font=\scriptsize\bfseries},
  io/.style={blk, fill=figgreen, draw=figgreenln},
  ours/.style={blk, fill=figorange, draw=figorangeln}]
  \node[io] (emb) {Embedding};
  \node[frz, above=of emb] (la)  {Layers 0--11 (frozen)};
  \node[inj, above=of la]  (l12) {Layer 12};
  \node[frz, above=of l12] (lb)  {Layers 13--15 (frozen)};
  \node[inj, above=of lb]  (l16) {Layer 16};
  \node[frz, above=of l16] (lc)  {Layer 17 (frozen)};
  \node[blk, fill=figgreen, draw=figgreenln, above=of lc] (l18) {Layer 18 (read)};
  \node[frz, above=of l18] (ld)  {Layer 19 (frozen)};
  \node[inj, above=of ld]  (l20) {Layer 20};
  \node[frz, above=of l20] (le)  {Layers 21--27 (frozen)};
  \node[io, above=of le]   (head) {RMSNorm + LM head};
  \foreach \a/\b in {emb/la, la/l12, l12/lb, lb/l16, l16/lc, lc/l18, l18/ld, ld/l20, l20/le, le/head} \draw[->] (\a) -- (\b);
  \node[below=2mm of emb, font=\scriptsize\itshape] (tok) {input tokens}; \draw[->] (tok) -- (emb);
  \node[ours, minimum width=52mm, right=22mm of l16, anchor=west, yshift=2mm] (wr)
    {\textbf{answer-only write} ($\times3$, Fig.~\ref{fig:steerarch} design)\\
     direction $\mathbf u(\mathbf r)$ $\cdot$ \textbf{self-muting magnitude} $a$\\
     {\tiny trained: answer CE (answerable rows only)}\\
     {\tiny $+$ magnitude suppression on \emph{unlabelled} OOD prompts}\\
     {\tiny $\Rightarrow$ in-domain $a$ active $\cdot$ under shift $a\!\to\!0$}};
  \node[blk, fill=figblue, draw=figblueln, minimum width=52mm, below=3mm of wr] (rh)
    {\textbf{supervised judge} (459K parameters)\\
     MLP $3584{\to}128{\to}1$ on the layer-18 state\\
     {\tiny trained: BCE on sufficiency labels, no gradient into the write}\\
     $\to$ insufficiency score $s$};
  \node[io, minimum width=52mm, above=3mm of wr] (dec)
    {\textbf{decision (label-free)}\\
     $s \geq \mathrm{Quantile}_{1-q}(\text{target scores})$ $\Rightarrow$ ``cannot answer''\\
     else\ answer $= \arg\max$ answer logits};
  \foreach \l/\dx in {l20/8.5mm, l16/11mm, l12/13.5mm}
    \draw[figblueln, dashed, ->] ([yshift=1.2mm]\l.east) -- ++(\dx,0) |- ([yshift=2mm]wr.west);
  \foreach \l/\dx in {l20/16mm, l16/18mm, l12/20mm}
    \draw[figorangeln, line width=0.9pt, ->] ([yshift=-2mm]wr.west) -- ++(-\dx,0) |- ([yshift=-1.2mm]\l.east);
  \draw[figblueln, dashed, ->] (l18.east) -- ++(7mm,0) |- (rh.west);
  \draw[->] (rh.east) -- ++(2.5mm,0) |- (dec.east);
  \draw[->] (head.east) -| ([xshift=-8mm]dec.west) -- (dec.west);
  \node[below=2.5mm of rh, font=\tiny\itshape, align=center, text=black!55]
    {one forward pass $\cdot$ backbone frozen $\cdot$ dashed = read (hook) $\cdot$ solid = write-back\\[1pt]
     operating points: in-domain-first (two-sided objective) $\cdot$ balanced ($\lambda{=}0.3$)};
\end{tikzpicture}
\caption{The supervised-judge variant, 7B instantiation (Qwen2.5-7B-Instruct: 28 layers,
hidden width 3584, write at layers 12/16/20, read at layer 18). Not part of YOPO. One pass,
write and read on separate channels: the answer-only write (orange) fires through a learned
magnitude trained to go to zero on unlabelled out-of-distribution prompts, and abstention
never touches the write. A small supervised judge (blue; an MLP trained with binary
cross-entropy, BCE, on sufficiency labels) scores sufficiency from the layer-18 state of the
same pass, and a label-free quantile threshold makes the abstain decision (green).}
\label{fig:yopo2}
\end{figure}

\section{The Four QA Datasets: All Rows at Three Scales}
\label{app:suite}
\begin{table}[h]\centering\scriptsize
\setlength{\tabcolsep}{4pt}
\caption{Answer-or-abstain accuracy on the four QA datasets at three scales, all rows
(Qwen2.5-Instruct 1.5B/3B/7B; test split, 2{,}000 items per dataset, half unanswerable; one
deterministic run per cell, no $\pm$). Rows per scale: the accuracy of the greedy answer on
the 1{,}000 answerable items alone (answer-correctness rule of App.~\ref{app:scoring});
generation alone (the prompt may reply ``unanswerable''); the two-pass reference (the
direction on a separate judgment-prompt pass); one pass (the direction on the generation
pass itself: no probe, no map, and a greedy generation identical to the generation-alone
row); cross-domain (the direction fit on MuSiQue-Full, threshold at the label-free median of
the target's own scores; ``(source)'' marks MuSiQue-Full itself, which has no cross-domain
cell); the oracle (gold-label abstention); and the direction's AUROC for answerable vs.\
unanswerable on the test split. On these datasets the direction's read layer is not the fixed
read layer of Table~\ref{tab:layers}: it is selected per dataset and scale as the layer with
the highest AUROC on the dataset's labelled selection split (its \emph{read layer}
in the table; 7B:
SQuAD2 20, RepLiQA 16, MuSiQue-C 17, MuSiQue-F 18; 3B: 26, 20, 23, 34; 1.5B: 27, 16, 16, 18;
the AUROC row is at that layer). The MuSiQue-Full direction transfers at 1.5B and 7B but not
at 3B, where the selection lands on layer 34 of 36, the second-to-last, and the cross-domain
cells fall below generation alone except MuSiQue-C ($0.588$ vs.\ $0.585$).}
\label{tab:taskc_scales}
\begin{tabular}{l l cccc}
\toprule
scale & system (answer-or-abstain accuracy unless noted) & SQuAD2 & RepLiQA & MuSiQue-C & MuSiQue-F\\
\midrule
1.5B & raw answerable-QA accuracy & $0.860$ & $0.571$ & $0.451$ & $0.156$\\
1.5B & generation alone (prompt may abstain) & $0.591$ & $0.439$ & $0.560$ & $0.410$\\
1.5B & two-pass reference (judgment-prompt pass) & $0.722$ & $0.767$ & $0.620$ & $0.458$\\
1.5B & \textbf{one pass}: direction on generation pass & $0.639$ & $0.754$ & $0.610$ & $0.462$\\
1.5B & cross-domain: direction fit on MuSiQue-Full & $0.726$ & $0.711$ & $0.609$ & (source)\\
1.5B & \textit{oracle (gold-label abstention)} & $0.930$ & $0.786$ & $0.726$ & $0.578$\\
1.5B & direction AUROC (test, read layer) & $0.842$ & $0.995$ & $0.833$ & $0.740$\\
\midrule
3B & raw answerable-QA accuracy & $0.836$ & $0.629$ & $0.273$ & $0.056$\\
3B & generation alone (prompt may abstain) & $0.672$ & $0.602$ & $0.585$ & $0.483$\\
3B & two-pass reference (judgment-prompt pass) & $0.763$ & $0.804$ & $0.598$ & $0.483$\\
3B & \textbf{one pass}: direction on generation pass & $0.748$ & $0.798$ & $0.592$ & $0.483$\\
3B & cross-domain: direction fit on MuSiQue-Full & $0.646$ & $0.567$ & $0.588$ & (source)\\
3B & \textit{oracle (gold-label abstention)} & $0.918$ & $0.815$ & $0.637$ & $0.528$\\
3B & direction AUROC (test, read layer) & $0.887$ & $0.999$ & $0.876$ & $0.732$\\
\midrule
7B & raw answerable-QA accuracy & $0.890$ & $0.632$ & $0.319$ & $0.072$\\
7B & generation alone (prompt may abstain) & $0.826$ & $0.622$ & $0.625$ & $0.508$\\
7B & two-pass reference (judgment-prompt pass) & $0.834$ & $0.805$ & $0.619$ & $0.508$\\
7B & \textbf{one pass}: direction on generation pass & $0.814$ & $0.797$ & $0.621$ & $0.508$\\
7B & cross-domain: direction fit on MuSiQue-Full & $0.827$ & $0.746$ & $0.624$ & (source)\\
7B & \textit{oracle (gold-label abstention)} & $0.945$ & $0.816$ & $0.660$ & $0.536$\\
7B & direction AUROC (test, read layer) & $0.926$ & $0.997$ & $0.870$ & $0.768$\\
\bottomrule
\end{tabular}
\end{table}

The \emph{supervised judge} of Table~\ref{tab:taskc_suite} is, per dataset, an MLP
($3584{\to}128{\to}1$) trained with binary cross-entropy (BCE) on the dataset's own labelled
fit split to score the layer-18 state of the judgment-prompt pass (probe off). It sees target
labels and is not part of YOPO; it is the in-domain comparison point for the label-free
direction.

\begin{table}[t]\centering
\small
\setlength{\tabcolsep}{5pt}
\caption{Answer-or-abstain accuracy on the four QA datasets, Qwen2.5-7B-Instruct, test split
(2{,}000 items per dataset), each gate OR-composed with the generation's own abstention
(App.~\ref{app:scoring}). The generation-alone, two-pass, cross-domain and oracle columns
repeat the 7B rows of Table~\ref{tab:taskc_main} (single deterministic runs, no $\pm$); the
supervised-judge column is the mean over three training seeds of the judge, and no seed
deviates from its mean by more than $0.002$. In-domain gates (the two-pass direction and the
judge) are fit and thresholded on the target's own fit and selection splits; the cross-domain
column fits the direction on MuSiQue-Full and sets the threshold label-free on the target;
``--'' marks MuSiQue-Full itself, the source, which has no cross-domain cell. Bold: the best
in-domain gate.}
\label{tab:taskc_suite}
\resizebox{\textwidth}{!}{%
\begin{tabular}{l c cc c c}
\toprule
 & & \multicolumn{2}{c}{in-domain gate} & cross-domain gate & \\
\cmidrule(lr){3-4}
target & generation alone & direction (two-pass) & supervised judge & direction fit on MuSiQue-Full & oracle \\
\midrule
SQuAD2    & $0.826$ & $0.834$ & $\mathbf{0.845}$ & $0.827$ & $0.945$ \\
RepLiQA   & $0.622$ & $0.805$ & $\mathbf{0.805}$ & $0.746$ & $0.816$ \\
MuSiQue-C & $0.625$ & $0.619$ & $\mathbf{0.642}$ & $0.624$ & $0.660$ \\
MuSiQue-F & $0.508$ & $0.508$ & $\mathbf{0.511}$ & --      & $0.536$ \\
\bottomrule
\end{tabular}}
\end{table}

\textbf{Per-dataset composition with the direction as the only gate.}
Table~\ref{tab:taskc_zs} shows how much of each dataset's gain comes from the generation's
own abstention and how much from the gate, against a calibrated judgment prompt and the
oracle (scoring rule: App.~\ref{app:scoring}). OR-composition is used throughout: the
generation's own abstention carries signal the gate does not, visibly on SQuAD2, where the
gate adds little over generation alone.

\begin{table}[h]\centering\small
\setlength{\tabcolsep}{3pt}
\caption{Answer-or-abstain accuracy with the direction as the only gate, Qwen2.5-7B-Instruct,
test split (2{,}000 items per dataset), single deterministic run, no $\pm$. Columns: accuracy
of the greedy answer on the 1{,}000 answerable items; answer-or-abstain accuracy for
generation alone, for generation OR a calibrated judgment prompt (a second pass in which the
frozen model is asked whether the context suffices; its yes/no logit margin is thresholded at
the label-free median of the selection split's margins), for generation OR the direction
(the two-pass row of Table~\ref{tab:taskc_main}), and for the oracle; last column, the
precision of the answers actually given (correct answers over items answered), generation
alone $\to$ with the direction gate. On RepLiQA the direction gate takes $0.183$ of the
$0.194$ gap between generation alone ($0.622$) and the oracle ($0.816$); the one-pass row of
Table~\ref{tab:taskc_main} takes $0.175$ of it.}
\label{tab:taskc_zs}
\resizebox{\textwidth}{!}{%
\begin{tabular}{l cc ccc c}
\toprule
 & answerable items & \multicolumn{4}{c}{answer-or-abstain accuracy} & answered-precision \\
\cmidrule(lr){3-6}
dataset & (raw accuracy) & generation alone & $+$calib.\ judgment prompt & $+$direction & oracle & gen.\ alone $\to$ $+$direction \\
\midrule
SQuAD2      & 0.890 & 0.826 & 0.830 & 0.834 & 0.945 & $0.749 \to 0.786$ \\
RepLiQA     & 0.632 & 0.622 & 0.748 & \textbf{0.805} & 0.816 & $0.459 \to \textbf{0.636}$ \\
MuSiQue-C   & 0.319 & 0.625 & 0.622 & 0.619 & 0.660 & $0.637 \to 0.645$ \\
MuSiQue-F   & 0.072 & 0.508 & 0.509 & 0.508 & 0.536 & $0.275 \to 0.275$ \\
\bottomrule
\end{tabular}}
\end{table}

\textbf{One-pass deployment.} The one-pass rows of Tables~\ref{tab:taskc_main} and
\ref{tab:taskc_scales} are a true one-pass product: no probe and no map run, the generation
is the frozen model's, and the direction reads the generation prompt's own hidden state at
the read layer before the first token is generated. The direction fit on the judgment-prompt
pass scores the generation-prompt state at $0.94$/$0.99$/$0.93$/$0.79$ AUROC on the selection
splits of SQuAD2/RepLiQA/MuSiQue-C/MuSiQue-Full at 7B (test-split values: the AUROC row of
Table~\ref{tab:taskc_scales}); the one-pass rows stay within $0.021$ of the two-pass rows at
7B and keep the RepLiQA gain at $+17.5\pp$ over generation alone. Whatever the model later
writes, chain-of-thought drafts included, cannot reach the gate. Two interference phenomena
are distinct here: a verdict written into the residual stream being overwritten by later
layers (write-to-speak), and the probe's write perturbing the direction's read
(write-to-read); this paper's map addresses only the second.

\textbf{Direction-only one pass (frozen generation).} Before any writer is trained for free-form
answers, the direction alone, read on the frozen model's own generation pass, already replaces
the judgment pass: Table~\ref{tab:taskc_main} (three scales) is within $0.021$ of the two-pass
reference in eleven of twelve cells; the exception is SQuAD2 on 1.5B ($0.639$ vs.\ $0.722$).
\begin{table}[h]\centering\scriptsize
\setlength{\tabcolsep}{3pt}
\caption{Direction only, no writer: answer-or-abstain accuracy on the frozen model's generation
pass (one pass) against the two-pass reference, generation alone and the oracle;
Qwen2.5-Instruct, test split, 2,000 items per dataset. The cross-domain row fits the direction on
MuSiQue-Full and sets the threshold on unlabelled target inputs.}
\label{tab:taskc_main}
\begin{tabular}{l l cccc}
\toprule
scale & system (answer-or-abstain accuracy) & SQuAD2 & RepLiQA & MuSiQue-C & MuSiQue-F\\
\midrule
1.5B & generation alone (prompt may abstain) & $0.591$ & $0.439$ & $0.560$ & $0.410$\\
1.5B & two-pass reference (judgment-prompt pass) & $0.722$ & $0.767$ & $0.620$ & $0.458$\\
1.5B & \textbf{one pass}: direction on generation pass & $0.639$ & $0.754$ & $0.610$ & $0.462$\\
1.5B & \textit{oracle (gold-label abstention)} & $0.930$ & $0.786$ & $0.726$ & $0.578$\\
\midrule
3B & generation alone (prompt may abstain) & $0.672$ & $0.602$ & $0.585$ & $0.483$\\
3B & two-pass reference (judgment-prompt pass) & $0.763$ & $0.804$ & $0.598$ & $0.483$\\
3B & \textbf{one pass}: direction on generation pass & $0.748$ & $0.798$ & $0.592$ & $0.483$\\
3B & \textit{oracle (gold-label abstention)} & $0.918$ & $0.815$ & $0.637$ & $0.528$\\
\midrule
7B & generation alone (prompt may abstain) & $0.826$ & $0.622$ & $0.625$ & $0.508$\\
7B & two-pass reference (judgment-prompt pass) & $0.834$ & $0.805$ & $0.619$ & $0.508$\\
7B & \textbf{one pass}: direction on generation pass & $0.814$ & $0.797$ & $0.621$ & $0.508$\\
7B & \textit{oracle (gold-label abstention)} & $0.945$ & $0.816$ & $0.660$ & $0.536$\\
7B & cross-domain: direction fit on MuSiQue-Full & $0.827$ & $0.746$ & $0.624$ & (source)\\
\bottomrule
\end{tabular}
\end{table}

\textbf{Writers on free-form QA: answerable accuracy and interference.}
Table~\ref{tab:m51interf} gives, for every (scale, writer, dataset) cell of
Table~\ref{tab:m51main}, the answerable-question accuracy of the frozen model and of the
writer-on generation (abstention ignored), and the direction's AUROC on the generation
prefill of the fit split with the writer off, on, and after the refit map. Writers are trained
on the answerable rows of the fit split only (SQuAD2 600, RepLiQA 225, MuSiQue-C 600,
MuSiQue-Full 600 items; three epochs, one for MuSiQue-Full) with the deployed generation prompt.
\begin{table}[h]\centering\scriptsize
\setlength{\tabcolsep}{3pt}
\caption{Answerable accuracy and the direction's AUROC on the generation prefill, per writer
(fit split for AUROC, test split for accuracy). Cells absent from the table were not run.}
\label{tab:m51interf}
\begin{tabular}{l l l cc c ccc}
\toprule
scale & writer & dataset & \multicolumn{2}{c}{answerable accuracy} & writer alone & \multicolumn{3}{c}{direction AUROC on the generation pass}\\
\cmidrule(lr){4-5}\cmidrule(lr){7-9}
 & & & frozen & writer on & (no abstention) & writer off & writer on & after map\\
\midrule
1.5B & probe & SQuAD2 & 0.860 & 0.867 & 0.448 & 0.874 & 0.839 & 0.865\\
1.5B & probe & RepLiQA & 0.571 & 0.626 & 0.322 & 0.994 & 0.992 & 0.993\\
1.5B & probe & MuSiQue-C & 0.451 & 0.625 & 0.313 & 0.883 & 0.850 & 0.881\\
1.5B & probe & MuSiQue-F & 0.156 & 0.300 & 0.150 & 0.780 & 0.745 & 0.778\\
1.5B & LoFiT & SQuAD2 & 0.860 & 0.889 & 0.449 & 0.874 & 0.817 & 0.861\\
1.5B & LoFiT & RepLiQA & 0.571 & 0.655 & 0.353 & 0.994 & 0.986 & 0.991\\
1.5B & LoFiT & MuSiQue-C & 0.451 & 0.612 & 0.306 & 0.883 & 0.794 & 0.891\\
1.5B & LoFiT & MuSiQue-F & 0.156 & 0.311 & 0.156 & 0.780 & 0.724 & 0.798\\
1.5B & LoReFT & SQuAD2 & 0.860 & 0.871 & 0.436 & 0.874 & 0.852 & 0.869\\
1.5B & LoReFT & RepLiQA & 0.571 & 0.649 & 0.326 & 0.994 & 0.986 & 0.992\\
1.5B & LoReFT & MuSiQue-C & 0.451 & 0.643 & 0.322 & 0.883 & 0.823 & 0.889\\
1.5B & LoReFT & MuSiQue-F & 0.156 & 0.319 & 0.160 & 0.780 & 0.700 & 0.762\\
\midrule
3B & probe & SQuAD2 & 0.836 & 0.903 & 0.479 & 0.915 & 0.885 & 0.915\\
3B & probe & RepLiQA & 0.629 & 0.727 & 0.419 & 0.997 & 0.997 & 0.997\\
3B & probe & MuSiQue-C & 0.273 & 0.717 & 0.367 & 0.911 & 0.896 & 0.917\\
3B & probe & MuSiQue-F & 0.056 & 0.382 & 0.191 & 0.776 & 0.718 & 0.780\\
3B & LoFiT & SQuAD2 & 0.836 & 0.904 & 0.473 & 0.915 & 0.870 & 0.915\\
3B & LoFiT & RepLiQA & 0.629 & 0.749 & 0.422 & 0.997 & 0.996 & 0.997\\
3B & LoFiT & MuSiQue-C & 0.273 & 0.707 & 0.354 & 0.911 & 0.867 & 0.918\\
3B & LoFiT & MuSiQue-F & 0.056 & 0.384 & 0.192 & 0.776 & 0.698 & 0.791\\
3B & LoReFT & SQuAD2 & 0.836 & 0.905 & 0.454 & 0.915 & 0.876 & 0.920\\
3B & LoReFT & RepLiQA & 0.629 & 0.721 & 0.362 & 0.997 & 0.991 & 0.991\\
3B & LoReFT & MuSiQue-C & 0.273 & 0.671 & 0.336 & 0.911 & 0.816 & 0.883\\
3B & LoReFT & MuSiQue-F & 0.056 & 0.409 & 0.205 & 0.776 & 0.654 & 0.789\\
\midrule
7B & probe & SQuAD2 & 0.890 & 0.938 & 0.530 & 0.947 & 0.931 & 0.947\\
7B & probe & RepLiQA & 0.632 & 0.735 & 0.468 & 0.998 & 0.996 & 0.998\\
7B & probe & MuSiQue-C & 0.319 & 0.756 & 0.393 & 0.902 & 0.885 & 0.907\\
7B & probe & MuSiQue-F & 0.072 & 0.451 & 0.227 & 0.802 & 0.787 & 0.803\\
7B & LoFiT & SQuAD2 & 0.890 & 0.947 & 0.552 & 0.947 & 0.937 & 0.948\\
7B & LoFiT & RepLiQA & 0.632 & 0.765 & 0.521 & 0.998 & 0.998 & 0.998\\
7B & LoFiT & MuSiQue-C & 0.319 & 0.733 & 0.370 & 0.902 & 0.898 & 0.907\\
7B & LoFiT & MuSiQue-F & 0.072 & 0.495 & 0.248 & 0.802 & 0.798 & 0.810\\
7B & LoReFT & SQuAD2 & 0.890 & 0.938 & 0.470 & 0.947 & 0.922 & 0.947\\
7B & LoReFT & RepLiQA & 0.632 & 0.751 & 0.377 & 0.998 & 0.998 & 0.999\\
7B & LoReFT & MuSiQue-C & 0.319 & 0.755 & 0.378 & 0.902 & 0.889 & 0.907\\
7B & LoReFT & MuSiQue-F & 0.072 & 0.501 & 0.251 & 0.802 & 0.762 & 0.809\\
\bottomrule
\end{tabular}
\end{table}

\textbf{Does the map move across datasets?} Table~\ref{tab:m51xmap} takes, for each source
dataset, its read layer, its direction and its map (all fit on that dataset, 1.5B probe) and
applies them to the other datasets' test sets, with each target's own writer on and each
target's own generations; the threshold is the label-free median on the target's sel split,
the rule of the cross-domain row above. Two things follow. First, on the two targets where the writer's loss is large, maps fit
on other datasets recover $53$--$70\%$ of it: on SQuAD2 the MuSiQue-Full map recovers $70\%$
($0.775\to0.806$ against $0.819$ with the writer off) and the MuSiQue-C map $53\%$; on MuSiQue-C
the MuSiQue-Full map recovers $53\%$ and the SQuAD2 map $57\%$. Where a direction barely reads
the target (the RepLiQA direction on every other dataset, AUROC $0.55$--$0.60$; the SQuAD2
direction on RepLiQA, $0.72$) its map recovers $5\%$ or less. Second, under the judgment-prompt read layer the SQuAD2 cell fell short because of the layer,
not the map: that layer (27) reads the generation prefill at $0.72$, whereas the MuSiQue-Full
direction at layer 18 reads the same prefill at $0.82$ and, deployed unchanged with its map,
gives $0.686$ answer-or-abstain accuracy against $0.562$ for the layer-27 configuration
(two-pass $0.711$). This is what motivated choosing the read layer on the deployed prompt
(next paragraph).
\begin{table}[h]\centering\scriptsize
\setlength{\tabcolsep}{3pt}
\caption{Cross-dataset map (1.5B probe, test split). Row: the source whose read layer,
direction and map are used. Column: the target dataset, with its own writer on during
generation. AUROC is unanswerable vs.\ answerable on the generation prefill with the writer
off, on, and after the source's map; accuracy is answer-or-abstain without and with the map,
label-free median threshold on the target's sel split. Diagonal cells are in-domain.}
\label{tab:m51xmap}
\resizebox{\textwidth}{!}{%
\begin{tabular}{ll cccc}
\toprule
source (layer) & & SQuAD2 & RepLiQA & MuSiQue-C & MuSiQue-F\\
\midrule
SQuAD2 (27) & AUROC off / on / map & $0.721\,/\,0.577\,/\,0.664$ & $0.720\,/\,0.685\,/\,0.686$ & $0.607\,/\,0.513\,/\,0.566$ & $0.531\,/\,0.497\,/\,0.551$\\
 & accuracy, no map $\to$ map & $0.516\to0.583$ & $0.521\to0.518$ & $0.407\to0.441$ & $0.311\to0.338$\\
\addlinespace
RepLiQA (16) & AUROC off / on / map & $0.547\,/\,0.553\,/\,0.549$ & $0.987\,/\,0.978\,/\,0.975$ & $0.598\,/\,0.572\,/\,0.567$ & $0.573\,/\,0.565\,/\,0.560$\\
 & accuracy, no map $\to$ map & $0.511\to0.510$ & $0.758\to0.752$ & $0.428\to0.422$ & $0.339\to0.338$\\
\addlinespace
MuSiQue-C (16) & AUROC off / on / map & $0.758\,/\,0.718\,/\,0.739$ & $0.700\,/\,0.810\,/\,0.811$ & $0.798\,/\,0.673\,/\,0.716$ & $0.622\,/\,0.567\,/\,0.586$\\
 & accuracy, no map $\to$ map & $0.620\to0.640$ & $0.620\to0.616$ & $0.533\to0.565$ & $0.389\to0.395$\\
\addlinespace
MuSiQue-F (18) & AUROC off / on / map & $0.819\,/\,0.775\,/\,0.806$ & $0.949\,/\,0.951\,/\,0.953$ & $0.808\,/\,0.735\,/\,0.773$ & $0.732\,/\,0.704\,/\,0.729$\\
 & accuracy, no map $\to$ map & $0.670\to0.686$ & $0.721\to0.719$ & $0.574\to0.595$ & $0.446\to0.451$\\
\bottomrule
\end{tabular}
}
\end{table}

\textbf{Choosing the read layer on the deployed prompt.} The read layer and direction of the
four-QA cells were first chosen on the judgment prompt (the protocol of the direction-only
rows above). Table~\ref{tab:m53} repeats
the 1.5B cells of all three writers with the layer and direction chosen on the generation prompt itself
(writer off; same rule: class-mean direction on the fit split, layer by sel-split AUROC), the
map, threshold and generations unchanged. SQuAD2 moves from layer 27 to 18 and YOPO from
$0.562$ to $0.697$ (two-pass $0.711$); MuSiQue-C from $0.570$ to $0.641$ ($0.638$); RepLiQA
from $0.761$ to $0.786$ ($0.792$); MuSiQue-Full keeps its layer and loses $0.017$. LoFiT and LoReFT
behave the same way (SQuAD2 $0.547\to0.690$ and $0.554\to0.691$; MuSiQue-C $0.593\to0.650$ and
$0.590\to0.661$, both above the reference), so with the layer chosen on the deployed prompt all
twelve 1.5B cells are within $0.03$ of the two-pass reference. The 1.5B shortfalls under the judgment-prompt layer were therefore the layer choice, not
the read or the map. Every cell of the main table uses this choice; it was registered after
Table~\ref{tab:m51xmap} and before any 3B or 7B cell was run under it.
\begin{table}[h]\centering\scriptsize
\setlength{\tabcolsep}{3pt}
\caption{Read layer chosen on the judgment prompt vs.\ on the generation
prompt, 1.5B, test split, answer-or-abstain accuracy. Prefill AUROC is on the fit split at the
generation-prompt layer, writer off / on / after the map. Both columns use the pre-registered
random-start map, so the generation-prompt values differ slightly from Table~\ref{tab:m51main},
which uses the identity start.}
\label{tab:m53}
\resizebox{\textwidth}{!}{%
\begin{tabular}{ll cc c ccc c}
\toprule
writer & dataset & \multicolumn{2}{c}{read layer} & prefill AUROC & \multicolumn{3}{c}{YOPO} & two-pass\\
\cmidrule(lr){3-4}\cmidrule(lr){6-8}
 & & judgment & generation & off / on / map & judgment layer & gen.\ layer, no map & gen.\ layer & \\
\midrule
probe & SQuAD2 & 27 & 18 & 0.874 / 0.839 / 0.866 & 0.562 & 0.665 & 0.697 & 0.711\\
probe & RepLiQA & 16 & 14 & 0.994 / 0.992 / 0.993 & 0.761 & 0.786 & 0.786 & 0.792\\
probe & MuSiQue-C & 16 & 17 & 0.883 / 0.850 / 0.876 & 0.570 & 0.622 & 0.641 & 0.638\\
probe & MuSiQue-F & 18 & 18 & 0.780 / 0.745 / 0.776 & 0.429 & 0.447 & 0.412 & 0.404\\
\midrule
LoFiT & SQuAD2 & 27 & 18 & 0.874 / 0.817 / 0.863 & 0.547 & 0.653 & 0.690 & 0.720\\
LoFiT & RepLiQA & 16 & 14 & 0.994 / 0.986 / 0.988 & 0.777 & 0.797 & 0.798 & 0.805\\
LoFiT & MuSiQue-C & 16 & 17 & 0.883 / 0.794 / 0.886 & 0.593 & 0.593 & 0.650 & 0.635\\
LoFiT & MuSiQue-F & 18 & 18 & 0.780 / 0.724 / 0.799 & 0.423 & 0.423 & 0.434 & 0.408\\
\midrule
LoReFT & SQuAD2 & 27 & 18 & 0.874 / 0.852 / 0.872 & 0.554 & 0.665 & 0.691 & 0.714\\
LoReFT & RepLiQA & 16 & 14 & 0.994 / 0.986 / 0.989 & 0.772 & 0.785 & 0.794 & 0.802\\
LoReFT & MuSiQue-C & 16 & 17 & 0.883 / 0.823 / 0.878 & 0.590 & 0.593 & 0.661 & 0.644\\
LoReFT & MuSiQue-F & 18 & 18 & 0.780 / 0.700 / 0.764 & 0.438 & 0.416 & 0.391 & 0.410\\
\bottomrule
\end{tabular}
}
\end{table}
\label{sec:arena}
\section{Scoring Rule for the Four QA Datasets}
\label{app:scoring}
Every answer-or-abstain number on SQuAD2, RepLiQA, MuSiQue-Curated and MuSiQue-Full is
produced by one deterministic scorer, released with the code; the one-pass deployment run
stores per-item generations and gate scores and is scored by the same rule. No LLM judge
enters any reported number.

\textbf{Generation.} One greedy pass per dataset, user prompt:
\begin{quote}\small\ttfamily
Passage: \{passage\}\\
Question: \{question\}\\
Answer concisely based only on the passage. If the passage does not contain enough
information to answer, reply "unanswerable".
\end{quote}

\textbf{Normalization} (SQuAD-official): lowercase, strip punctuation, delete the articles
\emph{a/an/the}, collapse whitespace.

\textbf{The answer-correctness rule.} An answerable item scores $1$ if and only if the system
does not abstain \emph{and} the generation matches the gold answer, where ``matches'' means:
the normalised gold string is contained in the normalised generation, or the token-level F1
between the two is at least $0.5$ ($\mathrm{F1}=2PR/(P{+}R)$ over multiset token overlap). An
unanswerable item scores $1$ if and only if the system abstains. Answer-or-abstain accuracy
is the mean of these per-item scores over the 2{,}000 test items of a dataset (1{,}000
answerable, 1{,}000 unanswerable); the ``answerable items (raw accuracy)'' rows apply the same
match rule to the 1{,}000 answerable items with abstention ignored.

\textbf{The abstention rule.} The generation itself abstains if and only if it contains the
string ``unanswerable'' (lower-cased), the reply the prompt asks for. A gated system abstains
if the generation abstains \emph{or} the gate scores the item insufficient (OR-composition).
Gate thresholds: in domain, the threshold that maximises accuracy on the ROC curve of the
dataset's labelled selection split (the split held out from the fit split,
App.~\ref{app:data}); cross-domain, the label-free median of the gate's scores on the
target's unlabelled selection split. The \emph{oracle} abstains exactly on the unanswerable
items.

A stricter cascade scorer ships with the replication study (containment in either direction
against any gold alias, an eleven-pattern abstention regex set, and a gray-zone file
($0.2<\mathrm{F1}<0.5$) as a hook for optional second-tier adjudication); no number reported
in this paper depends on it.

\section{Datasets and Checkpoints}
\label{app:data}
\textbf{Checkpoints.} All backbones are stock instruction-tuned checkpoints (Hugging Face
identifiers), frozen throughout; probes are the only trained parameters. Main scale
ladder \citep{qwen2025qwen25}: \texttt{Qwen/\allowbreak Qwen2.5-1.5B-Instruct}, \texttt{Qwen/\allowbreak Qwen2.5-3B-Instruct},
\texttt{Qwen/\allowbreak Qwen2.5-7B-Instruct}. Cross-family grid adds
\texttt{Qwen/\allowbreak Qwen2.5-0.5B-Instruct}, \texttt{allenai/\allowbreak OLMo-2-0425-1B-Instruct} \citep{olmo2025olmo2},
\texttt{TinyLlama/\allowbreak TinyLlama-1.1B-Chat-v1.0} \citep{zhang2024tinyllama}, \texttt{stabilityai/\allowbreak stablelm-2-1\_6b-chat} \citep{bellagente2024stablelm2},
\texttt{HuggingFaceTB/\allowbreak SmolLM2-1.7B-Instruct} \citep{allal2025smollm2}, \texttt{microsoft/\allowbreak Phi-3-mini-4k-instruct} \citep{abdin2024phi3},
and \texttt{mistralai/\allowbreak Mistral-7B-Instruct-v0.3} \citep{jiang2023mistral}. The four-QA-dataset generation runs use the three Qwen2.5-Instruct sizes with greedy
decoding.

\textbf{Per-backbone layers.} The probe's three injection layers sit at about $0.4$, $0.55$
and $0.7$ of the backbone's depth; Table~\ref{tab:layers} lists the layers and the exact
fractions for every backbone. On the Qwen 1.5B--7B sizes, OLMo-2, TinyLlama, StableLM-2 and
SmolLM2 the fractions lie within $0.04$ of $0.43/0.57/0.71$; the Qwen-0.5B, Phi-3-mini and
Mistral checkpoints use the layer sets listed, which sit within $0.1$ of them. The read
layer is the layer at which $\dsuff$ is fit and the single-layer map is trained (values
read from the released probe checkpoints and the pipeline scripts). On the four QA datasets
the direction's read layer is instead selected per dataset (App.~\ref{app:suite}).

\begin{table}[h]\centering\small
\caption{Injection and read layers for every backbone (0-indexed decoder blocks), with the
injection layers as fractions of depth. Probe hidden width 384, spread-5 positions. The
probes of the YOPO pipeline (every table of \S\ref{sec:yopo} and Apps.~\ref{app:family},
\ref{app:suite}, \ref{app:failures}) use $\alpha_{\max}{=}0.3$ with a learned per-position
magnitude; the steering grid of App.~\ref{app:steergrid} also ran $\alpha_{\max}{=}0.5$.}
\label{tab:layers}
\begin{tabular}{llcccc}
\toprule
backbone & size & depth & injection layers & injection layers / depth & read layer\\
\midrule
Qwen2.5 & 0.5B & 24 & \{8, 12, 16\} & 0.33 / 0.50 / 0.67 & 15\\
OLMo-2 & 1B & 16 & \{7, 9, 11\} & 0.44 / 0.56 / 0.69 & 10\\
TinyLlama & 1.1B & 22 & \{9, 13, 16\} & 0.41 / 0.59 / 0.73 & 14\\
Qwen2.5 & 1.5B & 28 & \{12, 16, 20\} & 0.43 / 0.57 / 0.71 & 18\\
StableLM-2 & 1.6B & 24 & \{10, 14, 17\} & 0.42 / 0.58 / 0.71 & 15\\
SmolLM2 & 1.7B & 24 & \{10, 14, 17\} & 0.42 / 0.58 / 0.71 & 15\\
Qwen2.5 & 3B & 36 & \{16, 20, 24\} & 0.44 / 0.56 / 0.67 & 24\\
Phi-3-mini & 3.8B & 32 & \{16, 20, 24\} & 0.50 / 0.63 / 0.75 & 20\\
Qwen2.5 & 7B & 28 & \{12, 16, 20\} & 0.43 / 0.57 / 0.71 & 18\\
Mistral & 7B & 32 & \{16, 20, 24\} & 0.50 / 0.63 / 0.75 & 20\\
\bottomrule
\end{tabular}
\end{table}

\paragraph{Write positions as implemented.} The five write positions are computed from the
attention mask of the left-padded batch. The released trainer and evaluator take the last real
token as index $n-1$ (with $n$ the prompt length), which is the last token only for the longest
row of a batch; for a padded row the ``last'' write moves earlier by the padding length or, when
the padding exceeds the prompt, into the padding, and the four spread positions shift with it.
Training pads to a multiple of 128, so on \anli{} the ``last'' write of about $45\%$ of training
rows landed in the padding (mean displacement 99 tokens); evaluation pads to the longest row in
a batch of four (mean displacement 8 tokens). Every \anli{} number in this paper was produced with that
code. Re-evaluating the 15k-item, 3-epoch 1.5B probe with the positions corrected (last real
token at $T-1$, spread positions from the true prompt length) changes \anli{} accuracy from
$0.728$ to $0.731$ (CPU, float32; the shipped bf16 value is $0.725$), and with batches of
sixteen the released evaluator gives $0.723$. A probe retrained with the corrected positions (5k items, 1 epoch) reaches $0.713$ against $0.706$
for the shipped checkpoint under the same evaluator. The corrected positions are used for the
free-form QA evaluation of \S\ref{sec:nativelabel} and are the default in the released code from
version 2 on; the checkpoints are unchanged.

\textbf{Datasets.} \emph{\anli{}} \citep{bhagavatula2020}: \texttt{allenai/art};
4-condition minimal-pair construction; 5{,}000 training items (seed 0), 1{,}532 held-out
validation items. \emph{HellaSwag} \citep{zellers2019hellaswag}: \texttt{Rowan/hellaswag},
binary reduction, 1{,}500-item transfer validation set. \emph{PIQA} \citep{bisk2020piqa}:
\texttt{ybisk/piqa} (steering grid). \emph{BoolQ} \citep{clark2019boolq}: validation
split, 1{,}499 items (plain accuracy in App.~\ref{app:lofit}; a full-vs-replace construction
in App.~\ref{app:failures}, v). \emph{The four QA datasets}: SQuAD2
\citep{rajpurkar2018}, RepLiQA \citep{monteiro2024repliqa} (\texttt{ServiceNow/repliqa}),
and MuSiQue \citep{trivedi2022musique} in two versions, MuSiQue-Curated (from its released
item list) and MuSiQue-Full (rebuilt deterministically from the validation split, seed 7,
with context windows that start with the supporting paragraphs so the gold chain is never
cut off). Test sets: 2{,}000 items per dataset, half unanswerable, verbatim from the
released item lists. Each dataset also has a labelled \emph{fit split}, on which the
direction (or a judge) is fit, and a \emph{selection split}, on which the read layer and the
in-domain threshold are chosen (and whose unlabelled scores set the cross-domain threshold):
SQuAD2, MuSiQue-Curated and MuSiQue-Full 1{,}200/800, RepLiQA 448/301; the reader of
\S\ref{sec:reader} uses 1{,}200/800 on every domain from its own 2{,}000-item pool (App.~\ref{app:readerscales}). The cross-matrix MuSiQue set
(App.~\ref{app:crossmatrix}) is a separate 2{,}000/1{,}000 balanced train/evaluation split
built the same way.

\subsection{Settings and Full Tables for \S\ref{sec:writer} and \S\ref{sec:yopo}}
\label{app:settings}
Tables~\ref{tab:writersets} and \ref{tab:fusedsets} list, for every comparison in the paper, the
backbone, the training split and the evaluation split, so that no number is read against the
wrong split. Table~\ref{tab:knobs} lists every fixed setting of YOPO, how it was chosen, and
what moves when it is changed. Table~\ref{tab:writers_full} is the full form of
Table~\ref{tab:writers}, including the probe's map applied to another writer without
refitting. Table~\ref{tab:combined_seeds} is the seed spread behind Table~\ref{tab:combined}.
Figure~\ref{fig:taskc} charts Table~\ref{tab:taskc_main}.

\begin{table}[h]\centering\scriptsize
\setlength{\tabcolsep}{4pt}
\caption{Backbone, training set and evaluation set for every writer comparison. ``Two-token
protocol'': the answer is read from the logits of the two option tokens. The headline $72.52$
(15k$\times$3 ep, $\alpha_{\max}{=}0.3$) and the appendix table's $72.5$ (5k$\times$1 ep,
$\alpha_{\max}{=}0.5$, the grid cell) are two different training runs that happen to coincide;
the probe that YOPO ships (\S\ref{sec:results}) is the 5k$\times$1 ep, $\alpha_{\max}{=}0.3$
configuration.}
\label{tab:writersets}
\begin{tabular}{p{3.0cm} p{2.7cm} p{3.2cm} p{3.7cm}}
\toprule
comparison & backbone(s) & trained on & evaluated on\\
\midrule
headline gain; competence-modulated series & Qwen2.5-Instruct 0.5B/1.5B/3B/7B; Phi-3-mini; Mistral-7B & \anli{}, 15k items, 3 ep, seed 0 (Mistral-7B: 5k, 1 ep) & \anli{} held-out validation, 1,532 items; two-token protocol; exact McNemar vs.\ the frozen baseline\\
matched-budget table; task-shift sentence & Qwen2.5-1.5B & \anli{}, 15k items, 3 ep, all three methods & \anli{} validation 1,532 (in-domain); zero-shot HellaSwag 1,500 and PIQA validation\\
injection geometry (App.~\ref{app:steergrid}) & Qwen2.5 1.5B/3B/7B & 5k items of the evaluated task, 1 ep, seed 0 (5k$\to$100k for the data-size point) & 1,532 held-out items of the same task: \anli{}, HellaSwag, PIQA\\
LoFiT / LoReFT with budget controls; writer-transfer round (App.~\ref{app:lofit}) & Qwen2.5 1.5B/3B/7B (transfer: 1.5B/7B) & \anli{} 5k, seed 0: probe 1 ep (and 3 ep); LoFiT 3 ep (and matched cost); LoReFT 3 ep & \anli{} validation 1,532; zero-shot HellaSwag 1,500, BoolQ 1,499\\
\bottomrule
\end{tabular}
\end{table}

\begin{table}[h]\centering\scriptsize
\setlength{\tabcolsep}{4pt}
\caption{What is fit on what, and where it is scored, for every evaluation of this section.
One recipe throughout; the map's weights are refit per (backbone, writer, seed) and never
transferred.}
\label{tab:fusedsets}
\begin{tabular}{p{2.6cm} p{2.3cm} p{3.7cm} p{3.6cm}}
\toprule
evaluation & backbone(s) & fit on & evaluated on\\
\midrule
AUROC loss and repair (\S\ref{sec:results}) & Qwen2.5-Instruct 1.5B / 3B / 7B & probe: \anli{} 5k, 1 ep, seed 0. $\dsuff$: the 600 full/replace-both rows of the \anli{} 4-cond train split; $M$: all 1,200 train rows, probe-off and probe-on residuals of the same passes & in-domain: \anli{} 4-cond val, 400 rows. HellaSwag 4-cond val, 500 rows, nothing fit on it. AUROC needs no threshold; for three-way accuracy the threshold is the 25th percentile of the evaluated set's own scores, on \anli{} and HellaSwag alike\\
other writers (\S\ref{sec:results}) & Qwen2.5 1.5B / 3B & $M$ refit on that writer's (probe-on, probe-off) \anli{}-train pairs; $\dsuff$ unchanged & as above (AUROC)\\
three-way accuracy; seeds (\S\ref{sec:benchmarks}) & 1.5B (4 seeds) / 3B / 7B (3 seeds each) & as above; $\dsuff$ and $M$ refit per seed & as above; abstention budget $p{=}25$\\
human labels, AUROC loss and repair (\S\ref{sec:nativelabel}) & Qwen2.5-1.5B / 3B & probe: the shipped \anli{} probe, unchanged. $\dsuff$ and $M$: the domain's fit split (SQuAD2 1,200; RepLiQA 448), judgment prompts & the dataset's 2,000-item test split, human answerable/unanswerable labels (AUROC)\\
human labels, deployment (\S\ref{sec:nativelabel}) & Qwen2.5-1.5B / 3B / 7B, greedy generation & writer (probe / LoFiT / LoReFT): the answerable rows of the dataset's fit split, free-form answer cross-entropy, deployed generation prompt; $M$: (writer-on, writer-off) prefill states of the fit split at the read layer; $\dsuff$ and the read layer: the generation-prompt fit split, layer by selection-split AUROC (App.~\ref{app:suite}); threshold: selection split, writer on. Direction-only rows (no writer): $\dsuff$ and threshold per dataset; cross-domain: $\dsuff$ fit on MuSiQue-Full only & test split, 2,000 items per dataset; two-pass reference = the same writer-on generation scored on a separate judgment-prompt pass\\
ten backbones (App.~\ref{app:family}) & Qwen2.5 0.5B--7B, OLMo-2, TinyLlama, StableLM-2, SmolLM2, Phi-3-mini, Mistral-7B & as the first row, refit per backbone; layers by proportional depth; nothing retuned & \anli{} 4-cond val; one seed per cell\\
\bottomrule
\end{tabular}
\end{table}

\begin{table}[h]\centering
\scriptsize
\setlength{\tabcolsep}{4pt}
\caption{Every fixed setting of YOPO. ``Fixed before the runs'' means the value was set
once and never searched; ``grid'' values come from the pre-registered steering grid.}
\label{tab:knobs}
\begin{tabular}{p{2.1cm} p{3.3cm} p{3.0cm} p{4.3cm}}
\toprule
setting & value & how chosen & what changes if moved\\
\midrule
injection layers & $\{12,16,20\}$ at 1.5B/7B, $\{16,20,24\}$ at 3B (about $0.4/0.55/0.7$ of depth; exact fractions in Table~\ref{tab:layers}) & grid over depth and layer count (App.~\ref{app:steergrid}) & layer count is a weak lever; the depth rule is what maps the probe to other backbones untouched\\
injection positions & five spread prompt positions & same grid & spread beats last-token by $5.6$ points on 1.5B (Table~\ref{tab:steergrid}, App.~\ref{app:steergrid}); three positions keep $94\%$ of the gain\\
write magnitude & $\alpha{=}0.3$ with a learned per-position magnitude & same grid & the learned magnitude is a null ($+0.26$ points, $p{=}0.767$)\\
probe size, data & hidden 384 ($0.23\%$ of backbone parameters on 1.5B, $0.11\%$ on 7B); 5k \anli{} items, 1 epoch & data sweep & 5k items, 1 epoch $\to$ 100k items, 3 epochs adds about $3$ points; saturates early\\
read layer & 18 at 1.5B/7B, 24 at 3B ($\approx0.65$ of depth) & fixed in advance, the layer at which the direction was first fit; not tuned here (on the four QA datasets the read layer is selected per dataset, App.~\ref{app:suite}) & where $\dsuff$ is fit and $M$ is applied\\
direction $\dsuff$ & mean(full) $-$ mean(replace-both) over the 600 full and replace-both rows of the 1{,}200-row training split, unit norm & by definition & no fitted weight; the only source-domain quantity in the read path\\
map $M$ & rank 64; $U$ initialised at zero (identity start), $V$ default; AdamW $10^{-3}$, weight decay $10^{-4}$; 400 full-batch steps; MSE & rank, optimiser and steps fixed before the runs; the identity start replaced the pre-registered random start after the four-QA 7B runs & the random start stalled on SmolLM2, Phi-3-mini and four 7B QA cells (App.~\ref{app:failures}, x); on \anli{} at 7B it had left the edited read in place (YOPO $0.880$), the identity start returns the read to the probe-off level (YOPO $0.858$, two-pass $0.863$)\\
abstain threshold & percentile $p{=}25$ of the evaluation set's own scores & matches the $25\%$ insufficient prior (one condition in four) & on \anli{} at 1.5B, $p{=}20/30$ move every policy by $\le1.5$ points and YOPO stays above each component at all three ($0.758/0.753/0.745$ at $p{=}20/25/30$); on HellaSwag at 1.5B the one-pass-no-map policy moves $2.6$ points at $p{=}30$ and YOPO at $p{=}20$ ($0.600$) is below the direction alone ($0.620$)\\
seeds & probe seed 0 in all tables; 4 seeds on 1.5B and 3 each on 3B and 7B in Table~\ref{tab:combined_seeds} & --- & in-domain std $\le0.016$, HellaSwag $\le0.010$\\
batching & batch 8, left padding to the batch's longest prompt & fixed & the write is sensitive to pad length (position ids); the released config reproduces every cell to four decimals\\
\bottomrule
\end{tabular}
\end{table}

\begin{table}[h]\centering\small
\setlength{\tabcolsep}{5pt}
\caption{Four writers, one map recipe (full form of Table~\ref{tab:writers}). AUROC of the same direction $\dsuff$ (full vs.\
replace) at the read layer on \anli{} (in-domain) and HellaSwag (transfer), Qwen2.5-1.5B/3B.
For each writer: the score on the pass with the writer on, and after a map refit on that
writer's own (on, off) \anli{}-train pairs. Last three rows: the probe's map applied to the
other writer's residuals without refitting. All maps are label-free; $\dsuff$ is fit once on
probe-off \anli{} train residuals. LoFiT and LoReFT are the official-configuration writers of
App.~\ref{app:lofit}, unchanged; LoRA is the matched-budget $r{=}3$ adapter of
Table~\ref{tab:matched} (its 3B counterpart trained identically).}
\label{tab:writers_full}
\resizebox{\textwidth}{!}{%
\begin{tabular}{ll cc cc}
\toprule
& & \multicolumn{2}{c}{1.5B} & \multicolumn{2}{c}{3B}\\
\cmidrule(lr){3-4}\cmidrule(lr){5-6}
writer & scored pass & \anli{} & HellaSwag (not fit on) & \anli{} & HellaSwag (not fit on)\\
\midrule
-- & two-pass reference (probe off) & $0.962$ & $0.919$ & $0.967$ & $0.922$\\
\midrule
probe & one pass, no map & $0.913$ & $0.836$ & $0.942$ & $0.891$\\
probe & one pass, map refit on this writer & $0.944$ & $0.888$ & $0.969$ & $0.908$\\
LoFiT & one pass, no map & $0.784$ & $0.751$ & $0.858$ & $0.807$\\
LoFiT & one pass, map refit on this writer & $0.947$ & $0.872$ & $0.942$ & $0.866$\\
LoReFT & one pass, no map & $0.730$ & $0.738$ & $0.939$ & $0.859$\\
LoReFT & one pass, map refit on this writer & $0.971$ & $0.892$ & $0.975$ & $0.880$\\
LoRA & one pass, no map & $0.811$ & $0.784$ & $0.972$ & $0.921$\\
LoRA & one pass, map refit on this writer & $0.962$ & $0.888$ & $0.974$ & $0.893$\\
\midrule
LoFiT & one pass, probe's map applied unchanged & $0.886$ & $0.812$ & $0.866$ & $0.842$\\
LoReFT & one pass, probe's map applied unchanged & $0.822$ & $0.791$ & $0.962$ & $0.874$\\
LoRA & one pass, probe's map applied unchanged & $0.895$ & $0.835$ & $0.980$ & $0.927$\\
\bottomrule
\end{tabular}
}
\end{table}

\begin{figure}[h]\centering
\includegraphics[width=\textwidth]{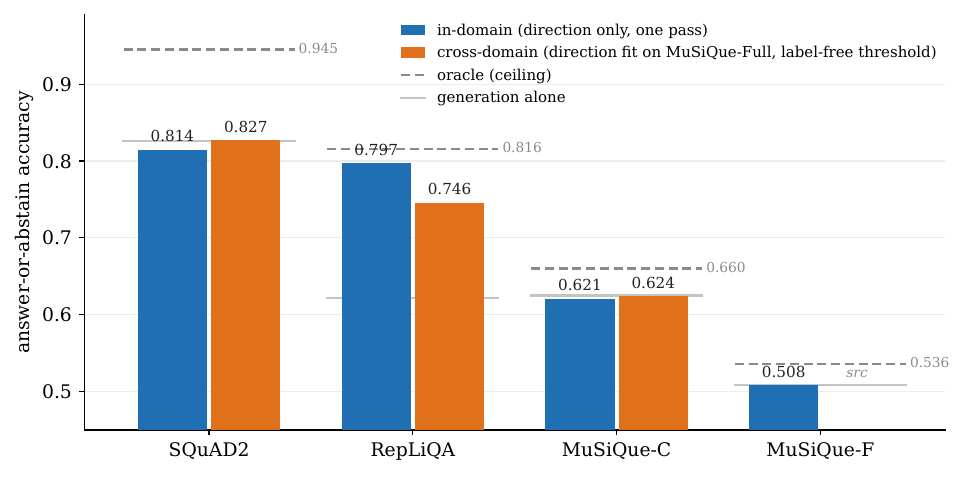}
\caption{Table~\ref{tab:taskc_main} at 7B as a chart (answer-or-abstain accuracy,
Qwen2.5-7B-Instruct, test split, 2{,}000 items per dataset). Blue: one pass, the direction on
the generation pass, in domain (no probe, no map, no labels). Orange: the direction fit on
MuSiQue-Full and deployed on the other three datasets with no target labels. The oracle
(dashed) and generation alone (solid gray; the prompt may abstain) bound each dataset; on the
multi-hop datasets the oracle nearly meets generation alone, so the limit there is answering,
not abstention.}
\label{fig:taskc}
\end{figure}

\textbf{Runs and changes after pre-registration.} The 3B seed runs, the 3B human-label runs, the LoRA
writer and the 1.5B/3B runs on the four QA datasets were added after the pre-registered set and
are reported without hypotheses. Two settings were changed after seeing results and both
settings are reported: on the four QA datasets the read layer is chosen on the generation
prompt rather than the judgment prompt (App.~\ref{app:suite}), and the map starts at the
identity rather than at PyTorch's random initialisation (App.~\ref{app:failures}, item (x),
Table~\ref{tab:m50}). The pre-registered seed check (the seed check: YOPO above both single
components on every seed) holds in domain at all three scales and fails on HellaSwag on 1.5B and
7B, where YOPO gains nothing over the direction alone (\S\ref{sec:benchmarks}).

\FloatBarrier
\begin{table}[h]\centering\small
\caption{Seed variance of the end-to-end benchmark (three-way accuracy, $p{=}25$; mean $\pm$
population std over 4 seeds at 1.5B and 3 seeds each at 3B and 7B, seed 0 included). \anli{} is in domain;
HellaSwag is not fit on.}
\label{tab:combined_seeds}
\resizebox{\textwidth}{!}{%
\begin{tabular}{lcccccc}
\toprule
policy (p=25) & \multicolumn{2}{c}{1.5B} & \multicolumn{2}{c}{3B} & \multicolumn{2}{c}{7B}\\
 & \anli{} & HellaSwag & \anli{} & HellaSwag & \anli{} & HellaSwag\\
\midrule
frozen model & $0.375\pm0.000$ & $0.500\pm0.000$ & $0.563\pm0.000$ & $0.400\pm0.000$ & $0.628\pm0.000$ & $0.604\pm0.000$\\
probe alone & $0.589\pm0.011$ & $0.497\pm0.003$ & $0.635\pm0.007$ & $0.517\pm0.012$ & $0.669\pm0.001$ & $0.590\pm0.002$\\
direction alone (probe off) & $0.560\pm0.000$ & $0.602\pm0.000$ & $0.743\pm0.000$ & $0.560\pm0.000$ & $0.828\pm0.000$ & $0.780\pm0.000$\\
two-pass reference & $0.754\pm0.010$ & $0.593\pm0.003$ & $0.802\pm0.008$ & $0.656\pm0.011$ & $0.863\pm0.001$ & $0.765\pm0.002$\\
one pass, no map & $0.717\pm0.013$ & $0.563\pm0.012$ & $0.777\pm0.014$ & $0.627\pm0.017$ & $0.871\pm0.007$ & $0.755\pm0.001$\\
\textbf{YOPO} (1 pass, label-free) & $0.754\pm0.016$ & $0.603\pm0.006$ & $0.805\pm0.004$ & $0.640\pm0.010$ & $0.868\pm0.007$ & $0.767\pm0.002$\\
\bottomrule
\end{tabular}
}
\end{table}

\section{Negative Results}
\label{app:failures}
Consistent with our pre-registration protocol, we report the variants that did not survive.
Unless stated otherwise, ``AUROC'' below is the direction's AUROC for full vs.\ replace-both
on the four-condition validation sets (\anli{}, 400 rows, in domain; HellaSwag, 500 rows,
nothing fit on it), Qwen2.5-Instruct, probe seed 0, exactly as in Table~\ref{tab:writers}.
$M_{\mathrm{ML}}$ denotes a multi-layer variant of the map, defined in (iii).

(i) \emph{Supervised MLPs on the probe-on read} (1.5B). In place of the label-free map, a
small MLP trained with BCE on the four-condition answerable-vs-not label scores the probe-on
read-layer state; two variants, one anchored to the direction (the direction's score plus a
learned correction) and one free. Both raise the in-domain AUROC on \anli{} to
$0.972$--$0.979$ (the map: $0.944$) but score $0.846$--$0.867$ on HellaSwag, below the map's
$0.888$. They see labels and transfer worse; not shipped.

(ii) \emph{A learned layer combination.} A learned weighting of the probe-on residuals of the
five layers $\{12,14,16,18,20\}$ (1.5B and 7B; $\{16,18,20,22,24\}$ at 3B), trained with BCE
on the four-condition label and read by the direction, scores HellaSwag at
$0.835$/$0.890$/$0.952$ AUROC at 1.5B/3B/7B against $0.836$/$0.891$/$0.952$ for the
uncorrected single-layer probe-on read: never better than doing nothing. Not shipped.

(iii) \emph{A multi-layer map $M_{\mathrm{ML}}$ does not converge at 7B.} $M_{\mathrm{ML}}$ is
an MLP ($5d\to512\to d$) that reconstructs the probe-off read-layer residual from the
concatenated probe-on residuals of the five layers of (ii), trained with the same MSE
objective as $M$ for 500 steps. At 1.5B and 3B it converges (\anli{}/HellaSwag AUROC
$0.959/0.869$ and $0.973/0.911$; the single-layer map: $0.944/0.888$ and $0.969/0.908$). At
7B it fits the mean and outputs a constant: AUROC $0.50$ in domain and on HellaSwag, with a
reconstruction MSE of $1.20$ after 500 steps and $0.67$ after $3{,}000$ (six times the
budget; $0.026$ at 1.5B): large-magnitude outlier channels of the 7B residual dominate the
raw MSE. A variant that standardises each input feature and centres the target repairs 7B
($0.983/0.967$) but lowers the HellaSwag AUROC at 1.5B ($0.869\to0.839$) and 3B
($0.911\to0.877$); its standardisation statistics are in-domain quantities and leak the task
signature by the very mechanism this paper studies. We therefore report the pre-registered recipe with its failure; the single-layer map is the scale-robust variant. The 7B failure
replicates on the BoolQ and RepLiQA constructions of (v) (AUROC $0.50$ and $0.07$); on
MuSiQue no number is reported.

(iv) At 3B a map trained with BCE on the four-condition label instead of the MSE objective
exceeds the probe-off AUROC on HellaSwag ($0.9324\pm0.0023$, sample std over five seeds,
against $0.922$ probe off and $0.908$ for the label-free map); at 7B it does not ($0.949$
against $0.968$ probe off, single seed). A scale-local observation, not a claim; it sees
labels.

(v) \emph{Our constructed cross-dataset sets fail as benchmarks.} We built full-vs-replace
sets on BoolQ and RepLiQA in the manner of \S\ref{sec:yopo} and scored the \anli{}-fit
direction on them through every path (probe off; probe on; the map; the map with a BCE
head; the BCE-trained map of iv): AUROC $0.98$--$1.00$ for every path at every scale,
saturated. We also built a harder RepLiQA set, full vs.\ ``answer sentence removed'' (the
same document with only the sentence containing the answer deleted; a different
construction from the four conditions of \S\ref{sec:yopo}): there every path, the probe-off
read included, floors at $0.56$--$0.68$ AUROC, and the map with a BCE head degrades most
(from the probe-on read's $0.669$ to $0.565$ at 1.5B). Neither set separates the paths; not
used.

(vi) \emph{Domain-adversarial training of the map does not help} (1.5B). We added a
gradient-reversal discriminator that tries to tell in-domain from HellaSwag probe-on
residuals at the map's output (unlabelled HellaSwag inputs only, no labels), with weight
$\lambda$. HellaSwag AUROC, mean over three seeds: $0.879$ at $\lambda{=}0$ (the plain map),
$0.877$ at $0.1$, $0.784$ at $0.3$, $0.833$ at $1.0$; probe off $0.919$. Neutral when weak,
harmful when strong: forcing domain invariance erases part of the sufficiency signal itself.
Not shipped.

(vii) \emph{A write-site relocation} ($\{12,16,20\}\to\{16,20,24\}$ on 7B) gained
$+3.13\pp$ in a free-generation measurement made outside this paper's protocol; under the
two-token \anli{} evaluation of \S\ref{sec:writer} (7B, frozen $77.6$) it loses $0.9$ and
$1.4\pp$ at $\alpha_{\max}{=}0.3$ and $0.5$. The gain is specific to the evaluation protocol;
we keep $\{12,16,20\}$.

(viii) \emph{No single supervised gate wins the four QA datasets both in and cross-domain}
(7B, answer-or-abstain accuracy, test split, 2{,}000 items per dataset). The supervised
judge of App.~\ref{app:suite} leads or ties every in-domain cell ($0.845/0.805/0.642/0.511$,
Table~\ref{tab:taskc_suite}) but, fit on one dataset and thresholded label-free on another,
falls to $0.644$ on RepLiQA from SQuAD2 and to $0.606$ on RepLiQA from MuSiQue-C, where the
direction fit on MuSiQue-Full gives $0.745$ in the same scoring run ($0.746$ in
Table~\ref{tab:taskc_main}). A judge trained on three datasets and tested on the fourth
reaches $0.797$/$0.687$/$0.625$ on SQuAD2/RepLiQA/MuSiQue-C, against $0.827$/$0.721$/$0.625$
for the direction fit on the same three. A variant that re-standardises the judge's input
with the target's own unlabelled statistics was also tried; no number is reported. Not
shipped.

(ix) \emph{A supervised gate on the one-pass read.} A BCE-trained head on the probe-on
residuals of the five layers of (ii) (standardised with in-domain statistics), added with a
learned weight to the direction's score through $M_{\mathrm{ML}}$, trained on the
four-condition answerable-vs-not label. It buys in-domain three-way accuracy on \anli{}
($0.798/0.830/0.893$ at 1.5B/3B/7B vs.\ YOPO's $0.753/0.800/0.858$) and beats the two-pass
reference in domain on all ten backbones of App.~\ref{app:family}, but pays on HellaSwag
($0.512$ vs.\ YOPO's $0.606$ at 1.5B; $0.696$ at 7B, below every abstaining system of
Table~\ref{tab:combined}), and at 7B its $M_{\mathrm{ML}}$ term is at chance (iii), so the
head alone acts. It sees labels; we do not ship it.

(x) \emph{The pre-registered map recipe (random start) stalls on two backbones and on 7B QA
cells; the identity start, used in every table, repairs them and costs $0.022$ on \anli{} at 7B.}
Under the pre-registered recipe (PyTorch default initialisation, 400 AdamW steps)
the reconstruction MSE on SmolLM2 and Phi-3-mini does not decrease at all, so the direction's
read is not repaired there ($0.823$ and $0.939$ AUROC, unchanged). A pre-registered check (pre-registration document in the \ifarxiv code repository\fi) re-fits the map with $U$
initialised at zero (an exact identity start) on the stored states of all ten backbones;
nothing is re-extracted. Table~\ref{tab:m50} reports it. The identity-initialised map moves
on every backbone and repairs the two failures (SmolLM2 $0.823\to0.914$, Phi-3
$0.939\to0.963$, full vs.\ replace-both); end to end it lifts SmolLM2 from $0.605$ to $0.755$
three-way accuracy (two-pass reference $0.778$) and StableLM-2 from $0.590$ to $0.618$. It is
not uniformly better: at 7B a converged map reconstructs the probe-off state, which reads
below the probe-on one ($0.979$ vs.\ $0.988$ on the definition of Table~\ref{tab:m50}), and
YOPO drops from $0.880$ to $0.858$ against the two-pass reference's $0.863$. The pre-registered recipe's 7B row is therefore an under-trained map that happens to preserve the probe-on
advantage, not the objective ``choosing'' the identity. Every table uses the identity start, except Table~\ref{tab:m53}, both of whose columns keep
the random start for a like-for-like comparison of read layers; the pre-registered numbers are in Table~\ref{tab:m50} and in this item. Of its two pre-registered hypotheses, the first (the identity-initialised map repairs the
stalled backbones) is supported; the second (it changes
the seven backbones the random start had already repaired by less than $0.01$ AUROC and $0.02$ three-way accuracy)
is not: StableLM-2 moves by $+0.024$ AUROC and $+0.028$ accuracy, TinyLlama by $+0.016$
AUROC ($0.841\to0.858$ in Table~\ref{tab:m50} after rounding), and Qwen-7B by $-0.022$
accuracy ($-0.0225$ before rounding). The AUROC values quoted for SmolLM2 and Phi-3-mini
above use the full vs.\ replace-both definition of Table~\ref{tab:family};
Table~\ref{tab:m50} uses the answerable vs.\ replace-both definition of the end-to-end
pipeline. For the two stalled backbones the identity-initialised map is the fix; on 7B on
\anli{} the random start had scored higher. On the free-form QA datasets of
\S\ref{sec:nativelabel} the 7B picture flips: there the probe (corrected write positions, App.~\ref{app:data}) does move the
read (AUROC $0.938\to0.848$ on SQuAD2, $0.792\to0.766$ on MuSiQue-Full), the pre-registered recipe again lowers the reconstruction error without moving the AUROC, and the
identity-initialised map (same stored states, 400 steps) restores it to the probe-off level
(Table~\ref{tab:zero7b}); scored end to end with the stored writer-on generations, it lifts
YOPO on SQuAD2 from $0.709$ to $0.802$ (two-pass $0.832$; both under the judgment-prompt read
layer, whereas Table~\ref{tab:m51main} uses the deployed-prompt layer and reaches $0.835$) and on MuSiQue-Full from $0.494$ to
$0.499$ ($0.509$), and leaves MuSiQue-C within $0.007$. The stall is a property of the random
start at 7B, not of the read, which is why the identity start is used throughout.
\begin{table}[h]\centering\scriptsize
\setlength{\tabcolsep}{3pt}
\caption{Registered vs.\ identity-initialised map on the 7B free-form QA cells (probe on,
corrected positions; same states, same direction). Fit-split and test AUROC are unanswerable
vs.\ answerable on the generation prefill; accuracy is answer-or-abstain on the stored
writer-on generations with the ROC-optimal threshold from the sel split.}
\label{tab:zero7b}
\resizebox{\textwidth}{!}{%
\begin{tabular}{l ccc cc ccc}
\toprule
& \multicolumn{3}{c}{fit-split AUROC} & \multicolumn{2}{c}{test AUROC} & \multicolumn{3}{c}{answer-or-abstain accuracy}\\
\cmidrule(lr){2-4}\cmidrule(lr){5-6}\cmidrule(lr){7-9}
dataset (7B, probe on) & probe off & registered map & identity-init map & registered & identity-init & YOPO registered & YOPO identity-init & two-pass\\
\midrule
SQuAD2 & 0.938 & 0.848 & 0.939 & 0.785 & 0.898 & 0.709 & 0.802 & 0.832\\
MuSiQue-C & 0.909 & 0.901 & 0.913 & 0.866 & 0.875 & 0.712 & 0.705 & 0.696\\
MuSiQue-F & 0.792 & 0.764 & 0.796 & 0.731 & 0.771 & 0.494 & 0.499 & 0.509\\
\bottomrule
\end{tabular}
}
\end{table}

\begin{table}[h]\centering\scriptsize
\setlength{\tabcolsep}{3pt}
\caption{Pre-registered map (random init, 400 steps) vs.\ identity-initialised map (400 and
1{,}600 steps), same stored states, same $\dsuff$; ten backbones, \anli{} four-condition
validation (400 rows) and HellaSwag (500 rows, nothing fit on it), single seed per cell, no
$\pm$. AUROC here is the end-to-end pipeline's definition, answerable (full, rm-obs1,
rm-obs2) vs.\ replace-both, not the full vs.\ replace-both of Table~\ref{tab:family}; YOPO is
three-way accuracy at $p{=}25$; ``two-pass'' is the two-pass reference. On HellaSwag only the
400-step maps were run, so there is no ``+4$\times$ steps'' column there.}
\label{tab:m50}
\resizebox{\textwidth}{!}{%
\begin{tabular}{l ccc ccc c cc}
\toprule
& \multicolumn{3}{c}{repaired-read AUROC (\anli{})} & \multicolumn{3}{c}{YOPO three-way (\anli{})} & & \multicolumn{2}{c}{YOPO (HellaSwag)}\\
\cmidrule(lr){2-4}\cmidrule(lr){5-7}\cmidrule(lr){9-10}
backbone & registered & identity-init & +4$\times$ steps & registered & identity-init & +4$\times$ steps & two-pass & registered & identity-init\\
\midrule
Qwen2.5 0.5B & 0.855 & 0.860 & 0.869 & 0.535 & 0.548 & 0.540 & 0.545 & 0.408 & 0.400\\
OLMo-2 1B & 0.924 & 0.929 & 0.933 & 0.645 & 0.638 & 0.638 & 0.635 & 0.514 & 0.508\\
TinyLlama 1.1B & 0.841 & 0.858 & 0.862 & 0.490 & 0.500 & 0.500 & 0.500 & 0.398 & 0.390\\
Qwen2.5 1.5B & 0.948 & 0.953 & 0.952 & 0.760 & 0.753 & 0.760 & 0.753 & 0.592 & 0.606\\
StableLM-2 1.6B & 0.903 & 0.927 & 0.926 & 0.590 & 0.618 & 0.620 & 0.625 & 0.512 & 0.502\\
SmolLM2 1.7B & 0.754 & 0.959 & 0.957 & 0.605 & 0.755 & 0.753 & 0.778 & 0.432 & 0.472\\
Qwen2.5 3B & 0.961 & 0.967 & 0.963 & 0.790 & 0.800 & 0.798 & 0.790 & 0.626 & 0.626\\
Phi-3-mini 3.8B & 0.931 & 0.962 & 0.962 & 0.575 & 0.575 & 0.578 & 0.578 & 0.540 & 0.552\\
Qwen2.5 7B & 0.988 & 0.979 & 0.978 & 0.880 & 0.858 & 0.855 & 0.863 & 0.754 & 0.768\\
Mistral 7B & 0.918 & 0.923 & 0.922 & 0.520 & 0.530 & 0.525 & 0.528 & 0.508 & 0.510\\
\bottomrule
\end{tabular}
}
\end{table}

\section{The Reader on Mistral-7B and at 1.5B and 3B}
\label{app:readerscales}
Tables~\ref{tab:reader_mistral}--\ref{tab:reader3b} repeat Table~\ref{tab:reader7b} on
Mistral-7B-Instruct-v0.3, Qwen2.5-1.5B-Instruct and Qwen2.5-3B-Instruct with the same protocol
(reader data seed 0). On Mistral-7B, LoRA is $24.6$ points above the baseline in domain and the
probe $11.3$; cross-domain they drop $24.6$ and $13.5$ points and end essentially level
with the baseline (LoRA $-0.01$, probe $-2.2$): the in-domain advantage that training
bought is erased entirely across domains. The read direction is $12.1$ points above the baseline
in domain and still $10.0$ above it cross-domain, a drop of only $2.1$ points. Qwen2.5-1.5B and
3B show the same structure: the read direction drops $3.4$ and $3.1$ points, and the two writers
(probe / LoRA) drop $16.7$ / $21.4$ and $18.0$ / $17.9$ points. On Mistral the writer
cells average three training seeds; at 1.5B and 3B each writer cell is a single training run
(seed 0).

\newcommand{\readertab}[6]{%
\begin{table}[h]\centering\scriptsize
\setlength{\tabcolsep}{2pt}
\caption{#2, same setup as Table~\ref{tab:reader7b}. Zero-shot baseline: #3.}
\label{#4}
\begin{tabular}{l cccc cccc cccc}
\toprule
& \multicolumn{4}{c}{read direction} & \multicolumn{4}{c}{probe} & \multicolumn{4}{c}{LoRA $r{=}3$}\\
\cmidrule(lr){2-5}\cmidrule(lr){6-9}\cmidrule(lr){10-13}
source $\backslash$ target & SQuAD2 & RepLiQA & MQ-C & MQ-F & SQuAD2 & RepLiQA & MQ-C & MQ-F & SQuAD2 & RepLiQA & MQ-C & MQ-F\\
\midrule
#1
\bottomrule
\end{tabular}
\end{table}}

\readertab{%
SQuAD2  & \textbf{.765} & .873 & .733 & .619 & \textbf{.764} & .694 & .608 & .559 & \textbf{.876} & .795 & .642 & .585\\
RepLiQA & .760 & \textbf{.906} & .689 & .622 & .723 & \textbf{.891} & .617 & .576 & .514 & \textbf{.981} & .499 & .500\\
MQ-C    & .773 & .828 & \textbf{.743} & .623 & .571 & .521 & \textbf{.718} & .595 & .645 & .600 & \textbf{.884} & .504\\
MQ-F    & .771 & .831 & .740 & \textbf{.624} & .654 & .544 & .735 & \textbf{.630} & .745 & .843 & .783 & \textbf{.795}\\
\midrule
in-domain mean & \multicolumn{4}{c}{.760} & \multicolumn{4}{c}{.751} & \multicolumn{4}{c}{.884}\\
cross-domain mean & \multicolumn{4}{c}{.738} & \multicolumn{4}{c}{.616} & \multicolumn{4}{c}{.638}\\
\begin{tabular}[t]{@{}l@{}}cross-domain $-$\\in-domain (points)\end{tabular} & \multicolumn{4}{c}{$-2.1$} & \multicolumn{4}{c}{$-13.5$} & \multicolumn{4}{c}{$-24.6$}\\
\begin{tabular}[t]{@{}l@{}}cross-domain $-$ prompting\\baseline (points)\end{tabular} & \multicolumn{4}{c}{$+10.0$} & \multicolumn{4}{c}{$-2.2$} & \multicolumn{4}{c}{$-0.0$}\\
}{Mistral-7B-Instruct-v0.3}{SQuAD2 .652, RepLiQA .536, MQ-C .735, MQ-F .630}{tab:reader_mistral}{}{}

\readertab{%
SQuAD2  & \textbf{.756} & .842 & .743 & .660 & \textbf{.755} & .674 & .617 & .523 & \textbf{.844} & .653 & .658 & .576\\
RepLiQA & .734 & \textbf{.955} & .728 & .668 & .699 & \textbf{.916} & .591 & .572 & .671 & \textbf{.994} & .646 & .570\\
MQ-C    & .754 & .849 & \textbf{.767} & .667 & .621 & .605 & \textbf{.804} & .544 & .617 & .644 & \textbf{.936} & .519\\
MQ-F    & .753 & .890 & .750 & \textbf{.671} & .610 & .607 & .687 & \textbf{.645} & .753 & .859 & .690 & \textbf{.700}\\
\midrule
in-domain mean & \multicolumn{4}{c}{.787} & \multicolumn{4}{c}{.780} & \multicolumn{4}{c}{.868}\\
cross-domain mean & \multicolumn{4}{c}{.753} & \multicolumn{4}{c}{.612} & \multicolumn{4}{c}{.654}\\
\begin{tabular}[t]{@{}l@{}}cross-domain $-$\\in-domain (points)\end{tabular} & \multicolumn{4}{c}{$-3.4$} & \multicolumn{4}{c}{$-16.7$} & \multicolumn{4}{c}{$-21.4$}\\
\begin{tabular}[t]{@{}l@{}}cross-domain $-$ prompting\\baseline (points)\end{tabular} & \multicolumn{4}{c}{$+12.5$} & \multicolumn{4}{c}{$-1.6$} & \multicolumn{4}{c}{$+2.7$}\\
}{Qwen2.5-1.5B-Instruct (writers: one training run, seed 0)}{SQuAD2 .710, RepLiQA .620, MQ-C .662, MQ-F .521}{tab:reader15b}{}{}

\readertab{%
SQuAD2  & \textbf{.800} & .859 & .753 & .657 & \textbf{.811} & .781 & .566 & .537 & \textbf{.862} & .707 & .680 & .599\\
RepLiQA & .774 & \textbf{.940} & .739 & .656 & .767 & \textbf{.974} & .603 & .562 & .780 & \textbf{.995} & .644 & .636\\
MQ-C    & .793 & .844 & \textbf{.774} & .665 & .537 & .669 & \textbf{.837} & .654 & .658 & .701 & \textbf{.924} & .647\\
MQ-F    & .790 & .860 & .762 & \textbf{.662} & .671 & .646 & .736 & \textbf{.674} & .776 & .804 & .737 & \textbf{.725}\\
\midrule
in-domain mean & \multicolumn{4}{c}{.794} & \multicolumn{4}{c}{.824} & \multicolumn{4}{c}{.876}\\
cross-domain mean & \multicolumn{4}{c}{.763} & \multicolumn{4}{c}{.644} & \multicolumn{4}{c}{.697}\\
\begin{tabular}[t]{@{}l@{}}cross-domain $-$\\in-domain (points)\end{tabular} & \multicolumn{4}{c}{$-3.1$} & \multicolumn{4}{c}{$-18.0$} & \multicolumn{4}{c}{$-17.9$}\\
\begin{tabular}[t]{@{}l@{}}cross-domain $-$ prompting\\baseline (points)\end{tabular} & \multicolumn{4}{c}{$+8.8$} & \multicolumn{4}{c}{$-3.1$} & \multicolumn{4}{c}{$+2.2$}\\
}{Qwen2.5-3B-Instruct (writers: one training run, seed 0)}{SQuAD2 .791, RepLiQA .774, MQ-C .590, MQ-F .545}{tab:reader3b}{}{}

\end{document}